\documentclass[10pt,twocolumn,letterpaper]{article}
\usepackage[accsupp]{axessibility}
\usepackage{cvpr}              
\usepackage{booktabs}  
\usepackage{multirow}  
\usepackage{makecell}  
\usepackage{xcolor}
\usepackage{bm}
\usepackage{fontawesome5}
\usepackage{bbding}
\usepackage{colortbl}
\usepackage{nicematrix}
\usepackage{url}

\newcommand{\dataset}{TASTE-Rob}
\definecolor{lightgreen}{RGB}{50,205,50}
\definecolor{deepgreen}{RGB}{21,142,73}   
\newcommand{\cmark}{\textcolor{deepgreen}{\CheckmarkBold}}
\definecolor{darkred}{RGB}{139,0,0}
\definecolor{lightgray}{RGB}{240,240,240}
\definecolor{lightblue}{RGB}{230,240,255}
\definecolor{lightorange}{RGB}{255,200,150}

\definecolor{cvprblue}{rgb}{0.21,0.49,0.74}
\usepackage[pagebackref,breaklinks,colorlinks,allcolors=cvprblue]{hyperref}

\def\paperID{3459} 
\def\confName{CVPR}
\def\confYear{2025}

\title{TASTE-Rob: Advancing Video Generation of Task-Oriented Hand-Object Interaction for Generalizable Robotic Manipulation}

\author{\linespread{0.1} \selectfont
Hongxiang Zhao$^1$\thanks{Equal contribution.}\hspace{1em}  
Xingchen Liu$^{1*}$\thanks{Work done as a visiting researcher at CUHKSZ.}\hspace{1em}  
Mutian Xu$^1$\hspace{1em}  
Yiming Hao$^1$\hspace{1em}  
Weikai Chen\thanks{This paper solely reflects the author's personal research and is not associated with the author's affiliated institution.}\hspace{1em}
Xiaoguang Han$^{1,2}$\thanks{Corresponding author: \href{mailto:hanxiaoguang@cuhk.edu.cn}{hanxiaoguang@cuhk.edu.cn}.}\\
\\
$^1$SSE, CUHKSZ\hspace{1em} $^2$FNii, CUHKSZ\\ 
\url{https://taste-rob.github.io}
}

\begin{document}
\maketitle

\begin{abstract}
We address key limitations in existing datasets and models for task-oriented {hand-object interaction} video generation, a critical {approach} of {generating video demonstrations for} robotic imitation learning. 
Current datasets, such as Ego4D~\cite{grauman2022ego4dworld3000hours}, often suffer from inconsistent view perspectives and misaligned interactions, leading to reduced video quality and limiting their applicability for precise imitation learning tasks.
Towards this end, we introduce TASTE-Rob — a pioneering large-scale dataset of 100,856 ego-centric hand-object interaction videos.
Each video is meticulously aligned with language instructions and recorded from a consistent camera viewpoint to ensure interaction clarity.
By fine-tuning a Video Diffusion Model (VDM) on TASTE-Rob, we achieve realistic object interactions, though we observed occasional inconsistencies in hand grasping postures. To enhance realism, we introduce a three-stage pose-refinement pipeline that improves hand posture accuracy in generated videos. 
Our curated dataset, coupled with the specialized pose-refinement framework, provides notable  performance gains in generating high-quality, task-oriented hand-object interaction videos, resulting in {achieving superior generalizable} robotic manipulation.
To foster further advancements in the field, TASTE-Rob dataset and source code will be made publicly available on our website \url{https://taste-rob.github.io}.
\end{abstract}

\section{Introduction}
\label{sec:intro}

Robotic manipulation plays an essential role in daily life, assisting with tasks like object handling, liquid pouring, surface cleaning, and drawer operation. Powered by the success of Imitation Learning (IL)~\cite{shridhar2022peract,chi2023diffusionpolicy,zhao2023learningfinegrainedbimanualmanipulation,rt12022arxiv,florence2021implicit,xu2023xskill,chanesane2023vip,jonnavittula2024viewvisualimitationlearning,jain2024vid2robotendtoendvideoconditionedpolicy,wang2023mimicplaylonghorizonimitationlearning,xiong2021learningwatchingphysicalimitation}, current robots can imitate actions from video demonstrations but require a nearly identical environment in the recorded video, limiting their generalization to novel scenes where factors like the position or movement direction of the manipulated object differ.

\begin{figure}[t]
  \centering
  \includegraphics[width=1.0\linewidth]{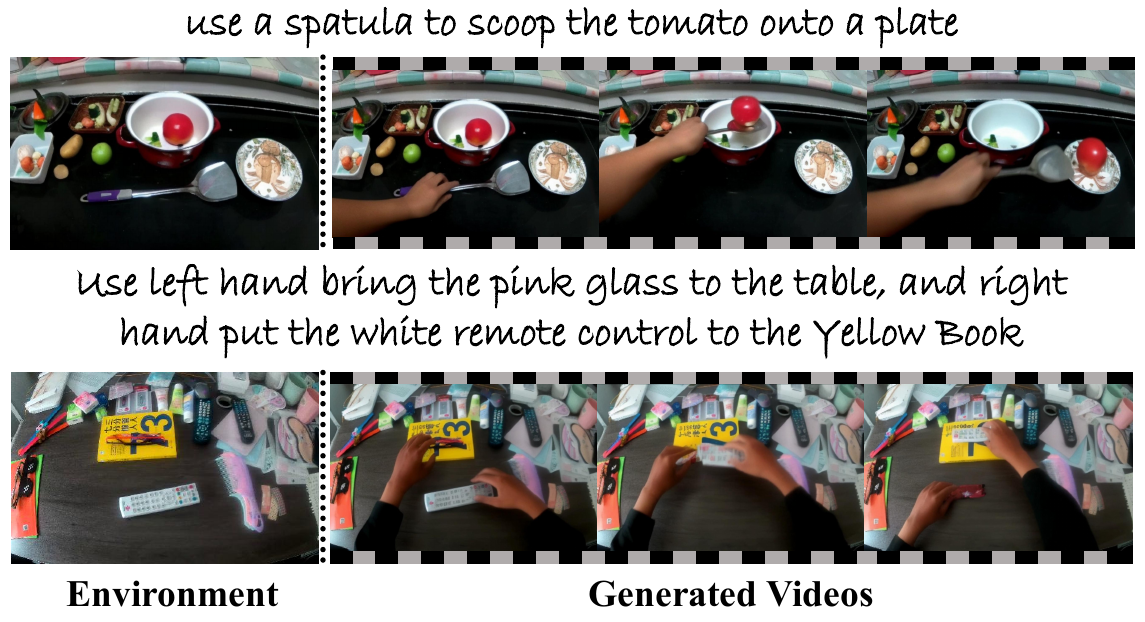}

\caption{\textbf{TASTE-Rob} generates high-quality task-oriented hand-object interaction videos, enabling robust generalization in robotic manipulation.}
   \label{fig:teaser}
   \vspace{-20pt}
\end{figure}


To solve the issue of limited generalization, recent research combines IL with generative models~\cite{du2023learninguniversalpoliciestextguided,wang2024thisthatlanguagegesturecontrolledvideo,black2023zeroshotroboticmanipulationpretrained,xu2024flow,BharadhwajVisual,bharadhwaj2023generalizablezeroshotmanipulationtranslating,bharadhwaj2024track2act,bharadhwaj2024gen2acthumanvideogeneration,ko2023learningactactionlessvideos}, like video diffusion models (VDMs), to create adaptable demonstrations and enable more generalizable robotic manipulation~\cite{du2023learninguniversalpoliciestextguided,wang2024thisthatlanguagegesturecontrolledvideo,ko2023learningactactionlessvideos}. {Specifically, these works develop VDMs to generate robot-object interaction videos as demonstrations for IL. 
While robot-object interaction VDMs enhance {generalization} performance, they are constrained by limited robot datasets~\cite{rt12022arxiv, bharadhwaj2023roboagentgeneralizationefficiencyrobot, jang2021bc, khazatsky2024droidlargescaleinthewildrobot, wang2024robotsonenewstandard}.} 
A scalable alternative is generating hand-object interaction (HOI) videos, which leverage extensive human manipulation video datasets.
Although hand-object interaction differs from robotic motion, advanced policy models~\cite{xu2023xskill,chanesane2023vip,jonnavittula2024viewvisualimitationlearning,zhu2024visionbasedmanipulationsinglehuman,wang2023mimicplaylonghorizonimitationlearning,bahl2022human,bharadhwaj2024gen2acthumanvideogeneration} can bridge this gap effectively, achieving remarkable performance through IL.
However, {even if existing powerful VDMs are able to generate remarkable human videos}, as shown in Figure \ref{fig:comp}, they struggle with generating high-quality, task-oriented hand-object interaction videos with accurate task understanding.

To enhance video generation quality, Ego4D~\cite{grauman2022ego4dworld3000hours}, a large-scale data containing 83,647 ego-centric hand-object interaction videos, has proven useful in robotic manipulation tasks~\cite{wu2023unleashing,ma2023livlanguageimagerepresentationsrewards,karamcheti2023voltron}. 
However, Ego4D presents two notable limitations for our purpose, as shown in Appendix \ref{appendix:dataset}. 
Firstly, the camera perspective in each clip is inconsistent, whereas video demonstrations for IL are typically recorded from a fixed viewpoint. 
Secondly, since these clips are extracted from longer videos that encompass multiple tasks, some clips contain partial interactions that overlap with others, resulting in misalignment between video clips and language instructions. 
Both limitations negatively impact the quality of video generation.

To address these above limitations, we introduce TASTE-Rob, a pioneering large-scale dataset designed to enhance the generation quality of task-oriented hand-object interaction videos.
TASTE-Rob contains 100,856 ego-centric videos with well-aligned language instructions.
Unlike Ego4D, TASTE-Rob is crafted to ensure clarity and consistency -- each video captures a single, complete interaction corresponding to its instruction and is recorded independently with a fixed camera viewpoint, mirroring the conditions needed for reliable imitation learning demonstrations.
By resolving these key issues, TASTE-Rob provides a robust foundation for advancing research in video generation, imitation learning, and related fields.

After fine-tuning a VDM on our curated dataset, we achieve high-quality video generation of task-oriented hand-object interactions. 
While the model demonstrates proficiency in identifying target objects and determining appropriate coarse interactions, it exhibits inconsistencies in hand poses, with the grip changing unnaturally during manipulation. This limitation presents a critical challenge for robotic imitation learning where precise and stable hand-object interactions are essential.
To address this issue, we further propose a dedicated three-stage pipeline for pose refinement. 
First, we generate a coarse manipulation video conditioned solely on language instructions and environment images.
Second, we refine the hand pose sequences using a Motion Diffusion Model (MDM)~\cite{tevet2023human}, ensuring realistic grip poses. 
Finally, we regenerate the videos by incorporating the revised pose sequences as an additional conditioning factor. 
This pipeline enhances grasping realism in generated videos, enabling robots to execute novel tasks with greater accuracy and adaptability in unseen scenarios.
Extensive experiments demonstrate that our proposed \dataset{} dataset, combined with the pose-refinement pipeline, achieves significant performance gains, surpassing state-of-the-art methods in the quality of generated hand-object interaction videos and achieving accurate robotic manipulation.

Our contributions can be summarized as follows:
\begin{itemize}
    \item We introduce \dataset{}, a large-scale dataset containing 100,856 ego-centric hand-object interaction videos, each specially tailored for training high-quality video generation model for robot imitation learning.
    \item We develop a three-stage pose-refinement pipeline that significantly enhances the fidelity of the hand posture in the generated videos.
    \item We set a new state of the art in generating task-oriented hand-object interaction videos, which further boosts the performance of robot manipulation.
\end{itemize}

\section{Related Works}

\paragraph{Imitation Learning from video demonstrations.} IL enables robots to learn from expert~\cite{shridhar2022peract,chi2023diffusionpolicy,zhao2023learningfinegrainedbimanualmanipulation,rt12022arxiv,florence2021implicit} or human~\cite{xu2023xskill,chanesane2023vip,jonnavittula2024viewvisualimitationlearning,jain2024vid2robotendtoendvideoconditionedpolicy,wang2023mimicplaylonghorizonimitationlearning,xiong2021learningwatchingphysicalimitation} video demonstrations by training policies to directly replicate expert actions through supervised learning. While these methods require video demonstrations from environments identical to the robot execution setup, they exhibit limited generalization to novel scenes and tasks. Alternatively, existing works address this challenge by generating demonstrations, such as videos~\cite{wu2023unleashing,ma2023livlanguageimagerepresentationsrewards,karamcheti2023voltron,bharadhwaj2024gen2acthumanvideogeneration}, optical flows~\cite{wen2024anypointtrajectorymodelingpolicy,xu2024flow}, and images~\cite{black2023zeroshotroboticmanipulationpretrained}, successfully achieving better generalization. Among existing approaches, video demonstration generation represents the most straightforward solution. Yet, robot manipulation video generation~\cite{wu2023unleashing,ma2023livlanguageimagerepresentationsrewards,karamcheti2023voltron} is constrained by limited robot datasets, while human hand video generation~\cite{bharadhwaj2024gen2acthumanvideogeneration} is hindered by the absence of effective HOI video generative models. In this work, we explore enhancing HOI video generation quality to facilitate robot manipulation.

\paragraph{Ego-Centric HOI Video datasets.} To enhance video quality, we propose a specialized HOI video generation model, which requires a large-scale, high-quality ego-centric dataset. Existing ego-centric HOI datasets have various limitations: insufficient resolution~\cite{6247820,6619194,li2020eyebeholdergazeactions,sigurdsson2018charadesegolargescaledatasetpaired}, limited scale~\cite{6247820,6619194,li2020eyebeholdergazeactions,sigurdsson2018charadesegolargescaledatasetpaired,6248010,6247805}, and restricted scene diversity~\cite{6247805,li2020eyebeholdergazeactions,Damen2022RESCALING}. While Ego4D~\cite{grauman2022ego4dworld3000hours} provides diverse, large-scale HOI videos, its varied camera perspectives and imprecise action-language alignment pose challenges for IL video demonstrations. Therefore, we introduce \dataset{}, containing 100,856 videos specifically designed for our task.

\paragraph{Diffusion Models for Video Generation.} Based on the success of VDMs~\cite{ho2022videodiffusionmodels}, researchers have achieved high-quality video generation with various conditioning inputs: text~\cite{mou2023t2iadapterlearningadaptersdig,shi2023instantboothpersonalizedtexttoimagegeneration,ye2023ipadaptertextcompatibleimage,zhang2023adding} and images~\cite{yang2024cogvideoxtexttovideodiffusionmodels,xing2023dynamicrafteranimatingopendomainimages,ren2024consisti2venhancingvisualconsistency,pku_yuan_lab_and_tuzhan_ai_etc_2024_10948109,xing2024tooncrafter}. Image-to-video generation aligns well with our task. While these models excel at producing realistic colors and temporal coherence, they struggle with significant motion variations. Motion diffusion models~\cite{tevet2023human,yang2023synthesizinglongtermhumanmotions,hu2023motionflowmatchinghuman,10376967,raab2024single} address this limitation by leveraging high-quality human body models~\cite{Romero_2017} to generate motion control parameters. In this work, we combine the advantages of VDM and MDM through our pose-refinement pipeline to generate HOI videos with realistic grasping poses.



\begin{table}
 \centering
 \begin{tabular}{@{}cccccc@{}}
   \toprule
   Orientation($^\circ$) & 0-90 & 90-180 & 180-270 & 270-360 \\
   \midrule
   Proportion($\%$) & 33.82 & 47.7 & 9.59 & 8.89 \\
   \bottomrule
 \end{tabular}
 \caption{\textbf{Palm orientation distribution of grasping poses.}
   We analyze the palm orientation distribution of grasping poses across \dataset{}. Specifically, the degree of orientation denotes the angle between the palm normal direction and the positive $x$-axis of the frame plane.}
 \label{tab:orient}
 \vspace{-10pt}
\end{table}

\section{Building \dataset{} Dataset}
\label{sec:dataset}

\begin{table*}
  \small
  \centering
  \begin{tabular}{@{}c|c|c|c|c|c|c|c@{}}
    \toprule
    \multirow{2}{*}[-0.5ex]{\centering Dataset} & \multicolumn{2}{c|}{Language Instructions} & 
    \multicolumn{2}{c|}{Environment} & \multicolumn{3}{c@{}}{Data Statistics} \\
    \cmidrule(l){2-3} \cmidrule(l){4-5} \cmidrule(l){6-8}
    & Perfect Action Alignment & Determiner & Camera & Setting & Frames & Clips & Resolution \\
    \midrule
    Disney~\cite{6247805} & \textcolor{darkred}{\XSolidBrush} & \textcolor{darkred}{\XSolidBrush} & Dynamic & Garden & 11M & 8800 & \cellcolor{lightorange}1080p\\
    ADL~\cite{6248010} & \textcolor{darkred}{\XSolidBrush} & \textcolor{darkred}{\XSolidBrush} & Dynamic & Diverse & 1M & 436 & 960p \\
    Charades-Ego~\cite{sigurdsson2018charadesegolargescaledatasetpaired} & \textcolor{darkred}{\XSolidBrush} & \textcolor{darkred}{\XSolidBrush} & Dynamic & Diverse & 5.8M & 7860 & 720p\\
    HOI4D~\cite{Liu_2022_CVPR} & \cmark & \textcolor{darkred}{\XSolidBrush} & Static & Diverse & 4M & 2466 & \cellcolor{lightorange}1080p \\
    UT Ego~\cite{6247820,6619194} & \textcolor{darkred}{\XSolidBrush} & \textcolor{darkred}{\XSolidBrush} & Dynamic & Diverse & 0.9M & 80 & 320p\\
    EGTEA Gaze+~\cite{li2020eyebeholdergazeactions} & \cmark & \textcolor{darkred}{\XSolidBrush} & Dynamic & Kitchen & 2.5M & 10,000 & 480p \\
    EPIC-KITCHENS~\cite{Damen2022RESCALING} & \cmark & \textcolor{darkred}{\XSolidBrush} & Dynamic & Kitchen & 6.8M & 35,000 & \cellcolor{lightorange}1080p \\
    Ego4D~\cite{grauman2022ego4dworld3000hours} & \textcolor{darkred}{\XSolidBrush} & \textcolor{darkred}{\XSolidBrush} & Dynamic & Diverse & \cellcolor{lightorange}19.2M & 83,647 & \cellcolor{lightorange}1080p \\
    \midrule
    \textbf{\textit{Ours}} & \cmark & \cmark & Static & Diverse & 9M & \cellcolor{lightorange}100,856 & \cellcolor{lightorange}1080p \\
    \bottomrule
  \end{tabular}
  \caption{\textbf{Dataset comparison with existing ego-centric HOI video datasets.} \dataset{}, specifically designed for generating task-oriented HOI videos, provides a diverse collection of ego-centric HOI videos with several distinctive features: static camera perspectives, precise language-action alignment, and rich object descriptors (Please see Appendix \ref{appendix:dataset} for details). In the above table, we compare these key characteristics: Perfect Action Alignment (ensuring all actions in the video strictly correspond to language task descriptions), Descriptive Determiners (whether object descriptions exist), Camera Setting (fixed or dynamic camera perspective), and Environment Setting (recording locations). By the way, Even if Ego4D~\cite{grauman2022ego4dworld3000hours} offers an impressive 3000-hours video data, it is designed for various general tasks, and the portion applicable to HOI video generation is significantly smaller, as presented in the above table, which is only 83,647 clips. Cells highlighted in \protect\scalebox{1}[0.6]{\protect\colorbox{lightorange}{\strut}} denotes the dataset with best feature in each column.}
  \label{tab:datasets}
  \vspace{-15pt}
\end{table*}

\begin{figure}[t]
  \centering
  \includegraphics[width=0.9\linewidth]{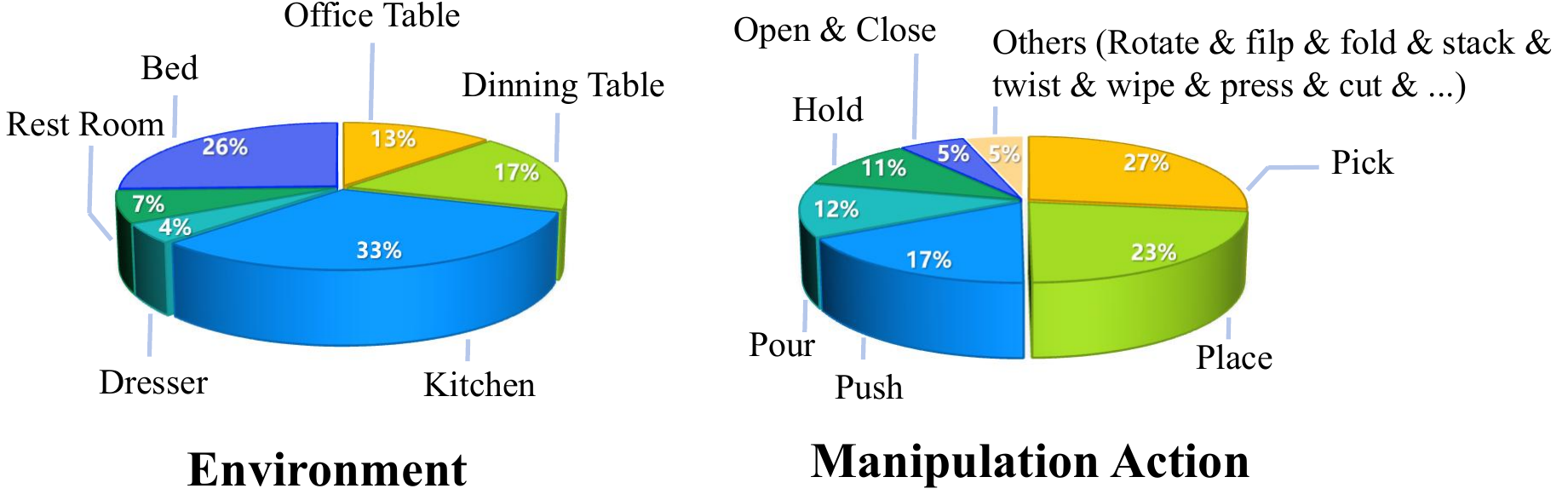}

   \caption{\textbf{Distribution of environment and HOI task.}
   In \dataset{}, we collect ego-centric task-oriented HOI videos with diverse tasks under diverse environments.}
   \label{fig:data_distribution}
   \vspace{-15pt}
\end{figure}

We collect a massive and diverse ego-centric task-oriented HOI video dataset \textbf{\dataset{}}, which includes 100,856 pairs of video and their corresponding language task instruction. To serve for HOI video generation, \dataset{} is required to achieve following goals: 1) Each video is recorded with a static camera perspective and contains a single action that closely aligns with the task instruction. 2) It encompasses diversified environments and tasks. 3) It features a variety of hand poses across different HOI scenarios.

\subsection{Collection Strategy and Camera Setting}

To achieve the first goal, we utilize multiple cameras equipped with a wide-angle lens capable of capturing 1080p ego-centric videos. We make the following improvements during each recording process.

Firstly, since our data collection aims to generate task-oriented HOI videos for demonstrations in IL, where demonstrations are typically recorded from fixed camera viewpoints for effective robot imitation learning, we ensure that no camera perspective variation occurs during recording. In addition, as shown in Figure~\ref{fig:example}, we align the camera perspective specifically to match the head-mounted camera setup of Ego4D~\cite{grauman2022ego4dworld3000hours}, ensuring consistency with the ego-centric perspective. 

Our second objective is to ensure precise alignment between language task instructions and video actions in \dataset{}, a crucial aspect for maintaining the action integrity in generated HOI videos. Unlike Ego4D~\cite{grauman2022ego4dworld3000hours}, which captures extended recordings of daily activities via head-mounted cameras and later segments them into  shorter clips, we employ a more controlled collection protocol: 1) Each video is strictly limited to under eight seconds in duration and captures a single action. 2) Collectors follow a structured recording process: they press the ``start recording" button, execute the specified HOI task according to the provided instruction, and stop recording upon task completion. This approach  ensures a precise correspondence between actions and task instructions.

\subsection{Data Diversity}




\paragraph{Distribution of environment and task.} To promote broad generalization, videos in \dataset{} are recorded across diverse environments and cover a wide range of HOI tasks. As shown in Figure \ref{fig:data_distribution}, environments include locations such as kitchens, bedrooms, dinning tables, office tables, etc. Collectors are required to interact with various commonly used objects, and perform tasks including picking, placing, pushing, pouring, etc. To further ensure task diversity, we consider different patterns of hand usage. Specifically, \dataset{} comprises 75,389 single-hand task videos and 25,467 double-hand task videos. 

\begin{figure}[t]
  \centering
  \includegraphics[width=0.8\linewidth]{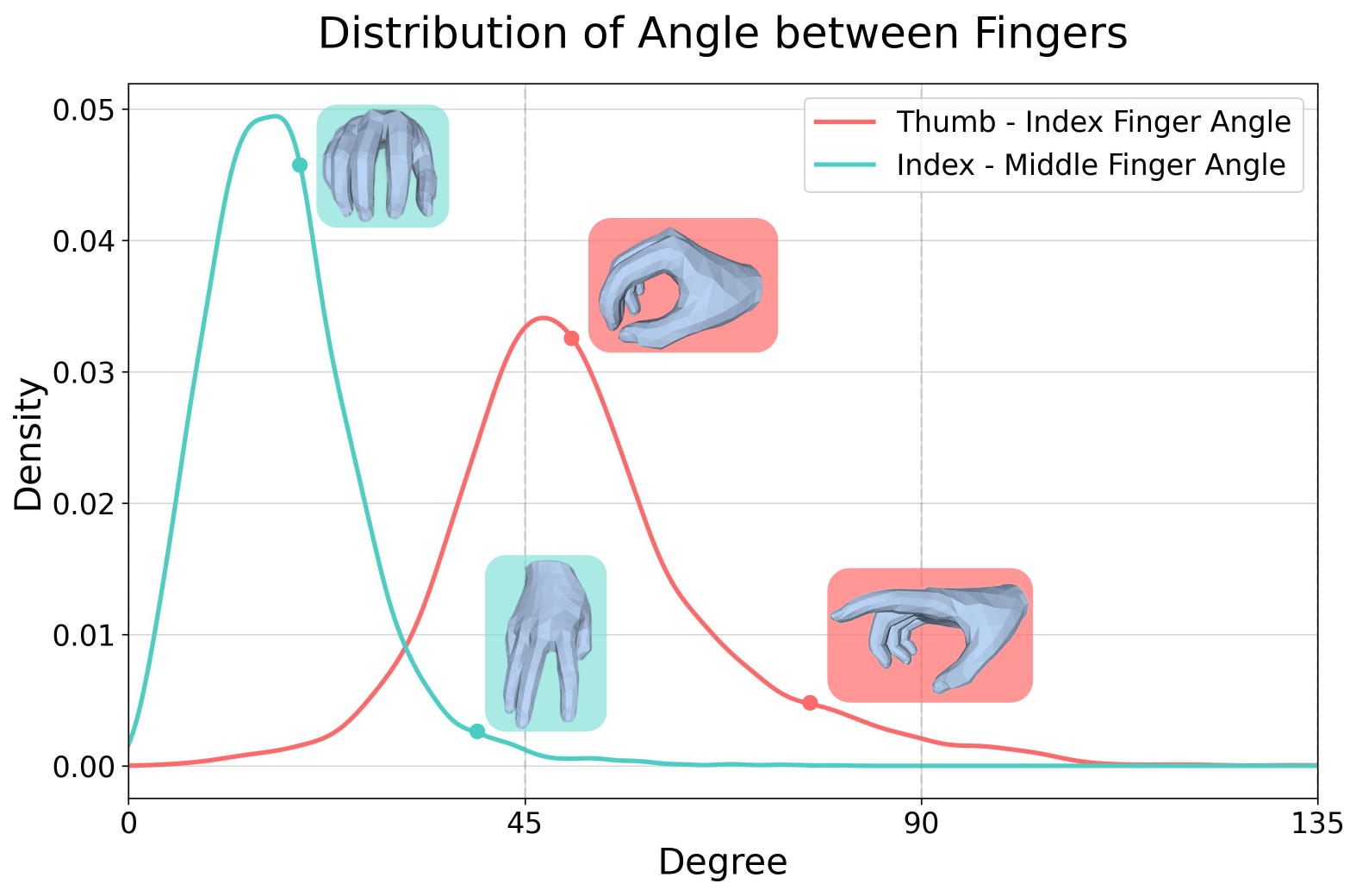}

   \caption{\textbf{Inter-finger angle distribution of human hands.}
   We analyze the distribution of inter-finger angles, specifically between the thumb-index and index-middle finger pairs, across \dataset{}. Specifically, higher degrees mean higher inter-finger angles.}
   \label{fig:angle}
   \vspace{-13pt}
\end{figure}

\begin{figure}[t]
  \centering
  \includegraphics[width=0.8\linewidth]{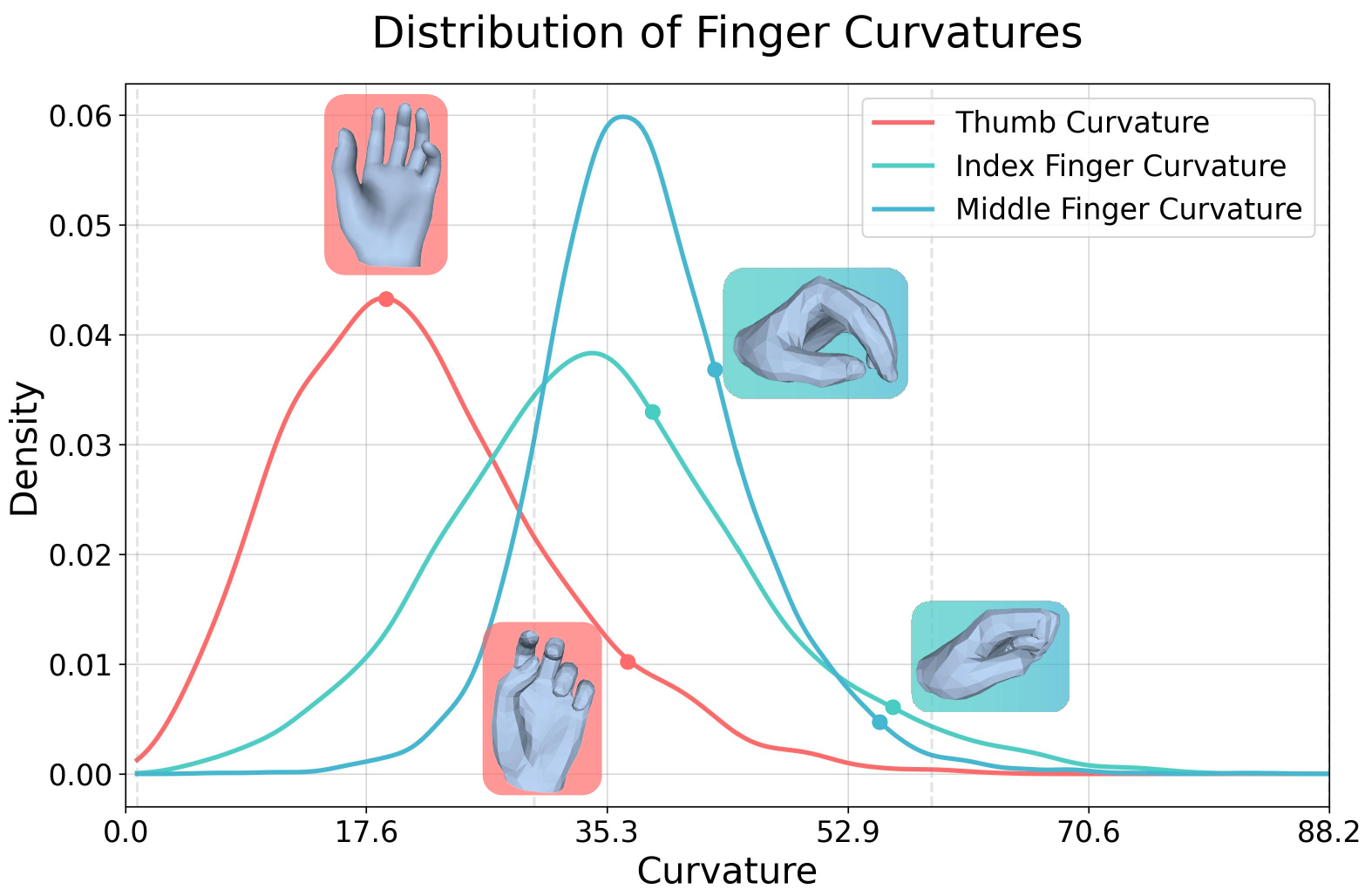}

   \caption{\textbf{Finger curvature distribution of human hands.}
   We analyze the distribution of finger curvatures, specifically that of thumb, index finger and middle finger, across \dataset{}. Specifically, higher degrees mean higher curvature.}
   \label{fig:bend}
   \vspace{-15pt}
\end{figure}

\paragraph{Distribution of Grasping Hands.} 
To ensure a wide variety of hand poses, we consider two primary factors: diverse palm orientations (global pose) and varied grasping poses (detailed pose). 
To demonstrate the diversity of hand poses, we utilize HaMeR~\cite{pavlakos2024reconstructing} to extract hand pose parameters and analyze the distribution based on these parameters (see further details in Appendix~\ref{appendix:hamer}). 

As shown in Table~\ref{tab:orient}, we analyze the distribution of palm orientations during HOI interaction across \dataset{}. The analysis reveals the following insights: 1)  Hand poses with palms facing down (0$^\circ$-180$^\circ$) are the most common, as this orientation is practical for grasping objects. 2) Hand poses with palms facing left (90$^\circ$-270$^\circ$) occur slightly more frequently than those facing right, likely because all collectors are right-handed and naturally prefer using their right hand for object manipulation.

We also provide the analysis of  hand grasping pose distribution in Figures~\ref{fig:angle} and \ref{fig:bend}. 
Given the dominant roles of the thumb, index finger, and middle finger in HOI, we focus on examining both the inter-finger angles between these fingers and their individual curvature distributions.
The analysis in Figure~\ref{fig:angle} shows a broad distribution of inter-finger angles, indicating diverse hand orientations. 
Figure \ref{fig:bend} reveals two key findings: 1) The curvature distributions of the index and middle fingers exhibit similar patterns, reflecting their synchronized bending during HOI actions. 2) Our dataset captures a broad spectrum of grasping poses, driven by the variety of objects being manipulated.

\subsection{Comparisons with Other Datasets}

As shown in Table~\ref{tab:datasets}, we provide comparison between \dataset{} and existing ego-centric HOI video datasets. \dataset{} is the first video dataset specifically designed for task-oriented HOI video generation, and it can also serve as a valuable resource for demonstrations in IL. 
Given that IL video demonstrations are recorded from a fixed camera perspective and include only a single action that aligns with the task instruction, we collect HOI videos under the same settings, distinguishing \dataset{} from other datasets.
In addition, to improve comprehension of target objects, we incorporate diverse determiners of objects in language task instructions (see Appendix \ref{appendix:dataset} for further details). With \dataset{}, we are able to generate high-quality HOI video demonstrations to achieve IL.

\section{Method}

Given an environment image and a task description, the generated task-oriented HOI videos are required to satisfy: 1) \textbf{Accurate Task Understanding}: correctly identifying which object to manipulate and how to manipulate it. 2) \textbf{Feasible HOI}: maintaining consistent hand grasping poses throughout the manipulation.

\begin{figure}[t]
  \centering
\includegraphics[width=0.9\linewidth]{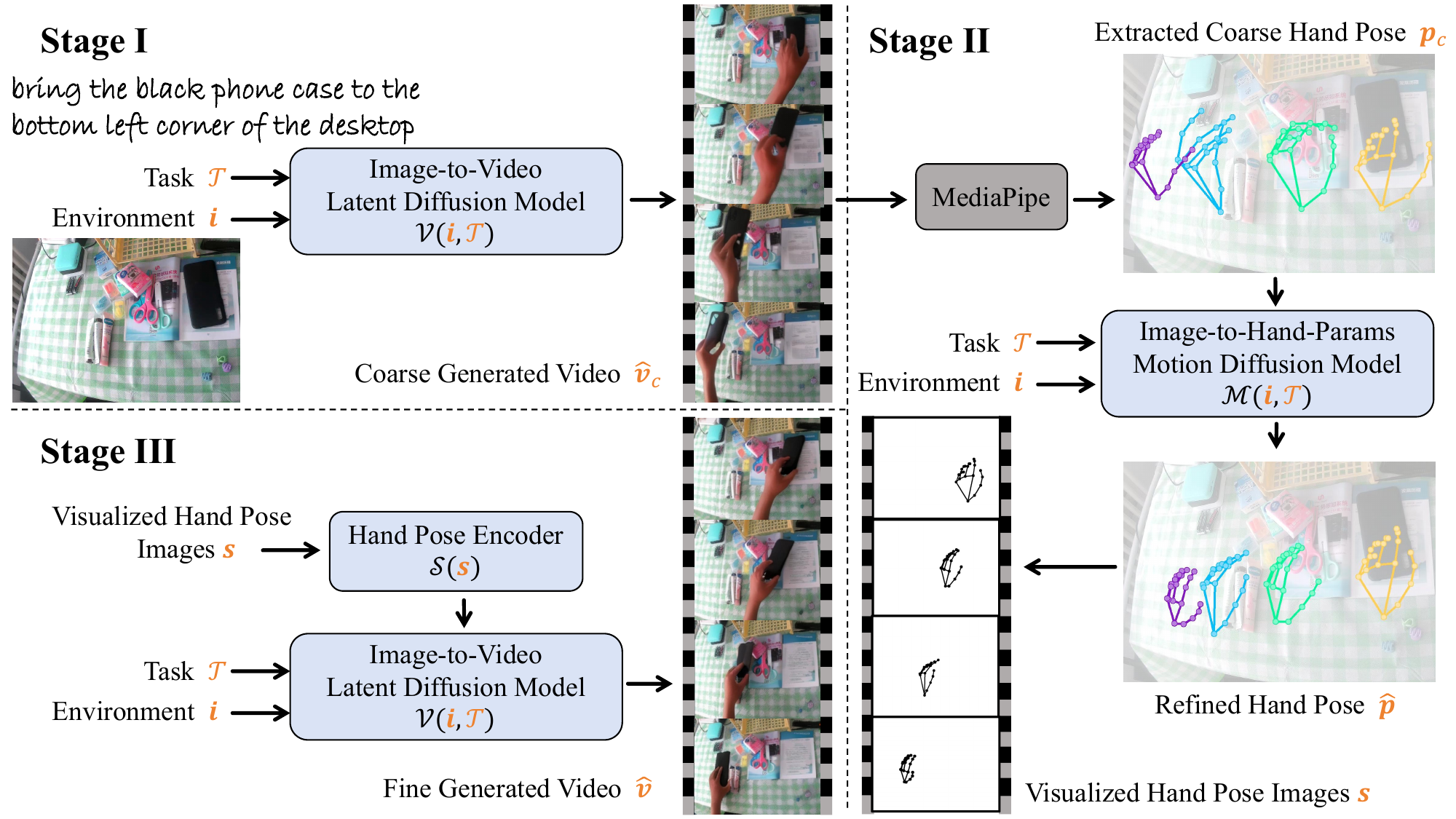}
\vspace{-5pt}
   \caption{\textbf{Overview of our proposed three-stage pose-refinement pipeline.} \textbf{Stage I}: our learnable I2V latent diffusion model $\mathcal{V}$ generates a coarse HOI video $\hat{\bm v}_c$. \textbf{Stage II}: extracting hand pose sequence $\bm p_c$ from $\hat{\bm v}_c$, and our learnable MDM $\mathcal{M}$ refines it into $\hat{\bm p}$. Besides, we also obtain the corresponding visualized hand pose images $\bm s$ of $\hat{\bm p}$. \textbf{Stage III}: incorporating this additional pose condition into $\mathcal{V}$ to generate a fine HOI video $\hat{\bm v}$.
   }
   \label{fig:framework}
   \vspace{-15pt}
\end{figure}

As illustrated in \textbf{Stage I} area of Figure \ref{fig:framework}, while videos generated by a single VDM ($\hat{\bm v}_c$) demonstrate accurate task understanding, exhibit limited fidelity in maintaining consistent grasping poses. To fulfill both requirements, we propose a three-stage pose-refinement pipeline, outlined in Figure \ref{fig:framework}: \textbf{Stage I}: employing a learnable Image-to-Video (I2V) diffusion model to generate a coarse HOI video that meets ``Accurate Task Understanding''. \textbf{Stage II}: extracting a hand pose sequence from this coarse video and refining it using a learnable Motion Diffusion Model (MDM)~\cite{tevet2023human}. \textbf{Stage III}: using the refined hand pose sequence to generate a high-fidelity HOI video that satisfies both requirements.

\subsection{Preliminary: Diffusion Models}

Diffusion models~\cite{ho2020denoisingdiffusionprobabilisticmodels,sohldickstein2015deepunsupervisedlearningusing} work by learning to reverse a noising process. They firstly transform data samples $\bm x_0 \sim p_{data}(\bm x)$  into Gaussian noise $\bm x_T \sim \mathcal{N} (\textbf{0},\textbf{I}) $, then learn to reconstruct the original data by denoising $\bm x_T$ with multiple steps. With the following objective:
\begin{equation}
  \min\limits_{\theta} \mathbb{E}_{t, \bm x, \epsilon \sim \mathcal{N} (\textbf{0},\textbf{I})} \Vert \epsilon - \epsilon_{\theta} (\bm x_t, t) \Vert_2^2, 
  \label{eq:ddpm}
\end{equation}
the model $\epsilon_{\theta} (\bm x_t, t)$ learns to predict single-step noise at each timestep $t$, where $\epsilon$ is the sampled ground truth noise and $\theta$ indicates the
learnable network parameters. After training, the model generates $\hat{\bm x}_0$ through a $T$-step denoising process from a random Gaussian noise $\bm x_T$.

\paragraph{Video Generation.} In this work, we explore HOI video generation based on DynamiCrafter~\cite{xing2023dynamicrafteranimatingopendomainimages}, a powerful I2V latent diffusion model. Suppose $\mathcal{T}$, $\bm i \in \mathbb{R}^{3 \times H \times W}$, and $\bm v \in \mathbb{R}^{L \times 3 \times H \times W}$ denote task language descriptions, environment images, and ground truth video frames, respectively. DynamiCrafter learns the denoising process in a compact latent space: $\bm{v}$ is encoded into a compact latent space through an encoder $\mathcal{E}$, yielding latent representations $\bm{z} = \mathcal{E}(\bm{v}) \in \mathbb{R}^{L \times C \times h \times w}$, and decoded via the decoder $\mathcal{D}$. Within this latent space, the model performs both forward diffusion and guided denoising processes conditioned on $\mathcal{T}$ and $\bm i$. Specifically, $\bm z_0 \sim p_{data}(\bm z)$ and $\bm z_T \sim \mathcal{N} (\textbf{0},\textbf{I})$. The optimization objective \ref{eq:ddpm} is updated as:
\begin{equation}
  \min\limits_{\theta_{V}} \mathbb{E}_{t, \bm z, \epsilon \sim \mathcal{N} (\textbf{0},\textbf{I})} \Vert \epsilon - \epsilon_{\theta} (\bm z_t, \mathcal{T}, \bm i, t, fps) \Vert_2^2, 
  \label{eq:i2v}
\end{equation}
where $fps$ is the FPS control introduced in \cite{xing2023dynamicrafteranimatingopendomainimages} and $\theta_{V}$ denotes the learnable parameters of DynamiCrafter.

\begin{figure}[h]
  \centering
  \includegraphics[width=0.8\linewidth]{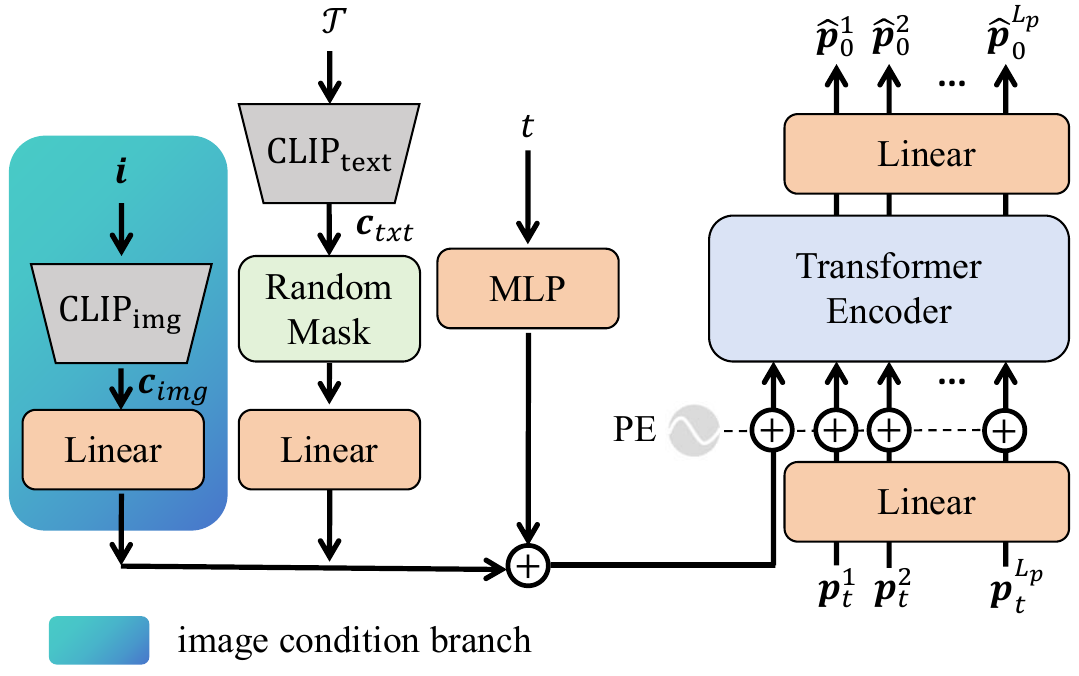}
    \vspace{-5pt}
   \caption{\textbf{overview of our learnable Image-to-Hand-Params MDM.} Based on MDM~\cite{tevet2023human}, we extend the framework by incorporating environment information through an additional image condition branch. The model takes a noised hand pose sequence $\bm p_t^{1:L_p}$, $t$ itself, a CLIP~\cite{radford2021learningtransferablevisualmodels} text embedding $\bm c_{txt}$ of $\mathcal{T}$ and a CLIP image embedding $\bm c_{img}$ of $\bm i$ as input.}
   \label{fig:imdm}
   \vspace{-15pt}
\end{figure}

\paragraph{Motion Sequence Generation.} MDM~\cite{tevet2023human}, utilizing a unique transformer encoder-only architecture, demonstrates remarkable performance in generating human motion sequences. 
Instead of predicting single-step noise, MDM directly generates clean motion sequences. Let $\bm p$ denotes motion sequences and $\mathcal{M}$ denotes MDM network. Here, $\bm p_0 \sim p_{data}(\bm p)$ and $\bm p_T \sim \mathcal{N} (\textbf{0},\textbf{I})$. The optimization objective \ref{eq:ddpm} is modified as:
\begin{equation}
  \min\limits_{\theta_M} \mathbb{E}_{t, \bm p, \epsilon \sim \mathcal{N} (\textbf{0},\textbf{I})} \Vert \bm p_0 - \mathcal{M} (\bm p_t, \mathcal{T}, t) \Vert_2^2, 
  \label{eq:mdm}
\end{equation}
where $\theta_M$ denotes the trainable parameters of MDM. After training, MDM generates the final clean motion sequence $\hat{\bm p}_0$ through a $T$-step denoising process. However, during each step, instead of directly generating $\bm p_{t-1}$  through single-step denoising, MDM firstly predicts the clean motion sequence $\hat{\bm p}_{0,t}$ from $\bm p_t$ with: 
\begin{equation}
  \hat{\bm p}_{0,t} = \mathcal{M} (\bm p_t, \mathcal{T}, t), 
  \label{eq:mdm-infer}
\end{equation}
and then reintroduces noises to obtain $\bm p_{t-1}$. Through repeating this process for $T$ times, MDM generates the final clean motion sequence $\hat{\bm p}_{0,0}$, denoted as $\hat{\bm p}_{0}$.

\subsection{Stage I: A Coarse Action Planner}

Our learnable coarse action planner is designed to generate a coarse HOI video $\hat{\bm v}_{c} \in \mathbb{R}^{L \times 3 \times H \times W}$ conditioned on task description $\mathcal{T}$ and environment image $\bm i$. Specifically, we fine-tune  DynamiCrafter~\cite{xing2023dynamicrafteranimatingopendomainimages} on \dataset{} to serve as the coarse action planner, denoted as  $\mathcal{V}$.


\paragraph{Training.} During fine-tuning, we adopt the training strategy similar to that of DynamiCrafter~\cite{xing2023dynamicrafteranimatingopendomainimages}. To leverage DynamiCrafter's robust temporal capabilities while adapting it to our specific HOI video generation, we only fine-tune its image context projector and spatial layers in its de-noising U-Net. The training objective remains consistent with Equation \ref{eq:i2v}, with trainable parameters $\theta_V$ representing parameters of the image context projector and spatial layers. 

\paragraph{Hand-Object Interacting Inconsistency Problem.} Our generated coarse HOI videos exhibit accurate task understanding, such as identifying the object to manipulate and determining its target position. 
However, as shown in $\bm p_c$ of Figure \ref{fig:framework}, the grasping poses exhibit  temporal inconsistencies during manipulation, indicating a lack of motion coherence. Specifically, these inconsistencies refer to undesirable variations in grasping poses over time, manifesting as unnatural changes in hand posture when the relative position between the hands and the grasped object should ideally remain stable. 
As shown in $\bm p_c$ of Figure \ref{fig:framework}, the \textcolor[RGB]{84,240,184}{green} hand pose demonstrates a pinching gesture, which is inconsistent with the grasping posture of the \textcolor[RGB]{253,207,87}{yellow} hand pose and unsuitable for the target object being manipulated.

\subsection{Stage II: Revising Hand Pose Sequences}

To address HOI inconsistencies, we train an Image-to-Hand-Params MDM $\mathcal{M}$. This model refines hand pose sequences $\bm p_c \in \mathbb{R}^{L_p \times N_h \times 2}$ extracted from the coarse video $\hat{\bm{v}}_c$. Specifically, we define $\bm p_c$ as the normalized coordinates of hand key-point sequences, where $L_p$ denotes the length of sequence and $N_h$ is the number of hand key-points. 

\paragraph{Training.} Our learnable $\mathcal{M}$ is designed to take the task description $\mathcal{T}$ and environment image $\bm i$ as input to predict refined human hand pose sequences. To achieve this, we extend the original MDM~\cite{tevet2023human} framework by incorporating environmental information through an additional image branch. 
As shown in Figure \ref{fig:imdm}, similar to the text condition branch of~\cite{tevet2023human}, the added image branch integrates CLIP~\cite{radford2021learningtransferablevisualmodels} class features of environment image $\bm i$. Consequently, the optimization objective \ref{eq:mdm} is updated as follows:

\begin{equation}
  \min\limits_{\theta_M} \mathbb{E}_{t, \bm p, \epsilon \sim \mathcal{N} (\textbf{0},\textbf{I})} \Vert \bm p_0 - \mathcal{M} (\bm p_t, \mathcal{T}, \bm i, t) \Vert_2^2, 
  \label{eq:i-mdm}
\end{equation}
where $\theta_M$ denotes the parameters of $\mathcal{M}$ and $\epsilon$
denotes the sampled ground truth noise. 

When we directly generate final clean pose sequence $\hat{\bm p}$ through a $T$-step denoising process, $\hat{\bm p}$ achieves physically plausible hand motions but exhibits limited spatial awareness. Conversely, $\bm p_c$ exhibits remarkable spatial awareness. 
To address this limitation, we refine $\bm p_c$ using $\mathcal{M}$ rather than generating from Gaussian noise. 
Specifically, we initialize the denoising process in Equation \ref{eq:mdm-infer} of $\mathcal{M}$ with $\bm p_{0,N_{rv}}$, setting $\bm p_c$ as $\bm p_{0,N_{rv}}$. Through an $N_{rv}$-step denoising, we refine $\bm p_c$ to obtain the final clean hand pose sequence $\hat{\bm p}$ which satisfies both spatial accuracy and motion feasibility.


\subsection{Stage III: Regenerating with Refined Pose}

With the refined hand pose sequences, we generate the fine-grained HOI videos with the additional pose condition $\hat{\bm p}$. 
Inspired by ToonCrafter~\cite{xing2024tooncrafter}, we train a frame-independent pose
encoder $\mathcal{S}$ to control the hand pose in the generated videos. We design $\mathcal{S}$ as a frame-wise adapter that adjusts the intermediate features of each frame independently, conditioned on $\hat{\bm p}$: ${\bm F}_{inject}^i = \mathcal{S}(\bm s^i, \bm z^i, t)$, where $\bm s^i$ is the visualized image sequences of $\hat{\bm p}$, shown as Figure \ref{fig:framework}, and ${\bm F}_{inject}^i$ is processed using a method similar to ControlNet~\cite{zhang2023adding}.

\paragraph{Training.} During training, we adopt a  strategy similar to ToonCrafter~\cite{xing2024tooncrafter}, where all parameters of $\mathcal{V}$ are frozen, and only the parameters of $\mathcal{S}$, denoted as $\eta$, are trained with the following optimization objective:
\begin{equation}
  \min\limits_{\eta} \mathbb{E}_{t, \bm s, \mathcal{E} (\bm v), \epsilon \sim \mathcal{N} (\textbf{0},\textbf{I})} \Vert \epsilon - \epsilon_{\theta}^{\mathcal{S}} (\bm z_t; \mathcal{T}, \bm i, \bm s, t, fps) \Vert_2^2, 
  \label{eq:ctrlnet}
\end{equation}
where $\epsilon_{\theta}^{\mathcal{S}}$ denotes the combination of $\epsilon_{\theta}$ and $\mathcal{S}$, and $\bm s$ represents the visualized images of hand pose sequences.

Finally, we consider our generated fine HOI videos $\hat{\bm v}$ as the video demonstrations for IL and enable robot manipulation using policy model of Im2Flow2Act~\cite{xu2024flow}. 
As illustrated in Figure \ref{fig:RM_case}, we present the imitation learning results with our generated HOI videos as demonstrations, confirming their effectiveness for enabling robot manipulation.

\section{Experiments}

\subsection{Implementation}

During training, we train our model on the training set of \dataset{} dataset. \textbf{Stage I}: We fine-tune our coarse action planner based on DynamiCrafter for 30K steps, with a batch size of 16 and a learning rate of $5 \times 10^{-5}$. \textbf{Stage II}: We train our MDM for 100K steps, with a batch size of 64 and a learning rate of $1 \times 10^{-4}$. \textbf{Stage III}: We fine-tune the pose encoder based on SD for 30K steps, with a batch size of 32 and a learning rate of $5 \times 10^{-5}$. During inference, we generate videos with a $50$-step denoising process and refine pose sequences with $N_{rv}$ as 10 (See more details in Appendix \ref{appendix:implementation}).

\subsection{Evaluation}
\paragraph{Datasets.} We evaluate on two datasets: \dataset{}-Test and Mujoco simulation dataset. \textbf{\dataset{}-Test}: We split \dataset{} dataset into training and test sets, with the test set denoted as \dataset{}-Test. For fair evaluation, we select $2\%$ of samples from each action category as test samples. \textbf{Mujoco simulation dataset}: we randomly select 50 tasks from robot simulation platform provided by Im2Flow2Act~\cite{xu2024flow} for robot manipulation.

In addtion, as mentioned in Section \ref{sec:dataset} and Table \ref{tab:datasets}, other ego-centric video datasets feature different settings from our task, making them unsuitable as evaluation datasets.

\paragraph{Metrics.} We evaluate the performance on three aspects: video generation, grasping pose consistency and robot manipulation. \textbf{Video Generation}: To evaluate the quality of generated videos in both the spatial
and temporal domains, we report Fr\'{e}chet Video Distance
(FVD)~\cite{Unterthiner2019FVDAN}, Kernel Video Distance (KVD)~\cite{Unterthiner2019FVDAN} and Perceptual Input Conformity (PIC)~\cite{xing2023dynamicrafteranimatingopendomainimages} as used in DynamiCrafter~\cite{xing2023dynamicrafteranimatingopendomainimages}. \textbf{Grasping Pose Consistency}: We evaluate with three metrics -  \textbf{Grasping Pose Variance} (GPV, computed by $\frac{1}{L^{'}}\sum_{i=1}^N(\theta^p_i-\bar{\theta^p})^2$), \textbf{Grasp Type Classification Error} (GTCE, calculating the classification error of grasping types in generated videos) and \textbf{Hand Movement Direction Accuracy} (HMDA, Measuring the directional consistency of hand movements and their alignment with the intended task trajectory),  
 and details of these metrics are illustrated in Appendix \ref{appendix:GPV}. \textbf{Robot Manipulation }: We utilize videos generated on the Mujoco simulation dataset as demonstrations for IL, and deploy the policy model of Im2Flow2Act~\cite{xu2024flow} for robot manipulation. The performance is evaluated by comparing the success rates.

\begin{table}[h]
 \centering
 \vspace{-5pt}
 \begin{tabular}{@{}cccc@{}}
   \toprule
   \multirow{2}{*}[-0.5ex]{\centering Method} & \multicolumn{3}{c}{Generation Quality} \\
   \cmidrule(l){2-4}
   & KVD$\downarrow$ & FVD$\downarrow$ & PIC$\uparrow$ \\
   \midrule
   consistI2V & 0.24 & 65.77 & 0.85 \\
   DynamiCrafter & 0.21 & 62.36 & 0.79 \\
   Open-Sora Plan & 0.18 & 50.19 & 0.84 \\
   CogVideoX & 0.16 & 48.72 & 0.85 \\
   \midrule
   \textbf{TASTE-Rob} & \textbf{0.03} & \textbf{9.43} & \textbf{0.90} \\
   \bottomrule
 \end{tabular}
 \vspace{-5pt}
 \caption{\textbf{Quantitative comparison between TASTE-Rob and baselines.} All video generation metrics prove the better generation performance of TASTE-Rob.}
 \label{tab:vid_comp_baselines}
 \vspace{-15pt}
\end{table}

\begin{figure*}[h]
  \centering
  \includegraphics[width=1.0\linewidth]{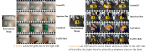}
    \vspace{-5pt}
\caption{\textbf{Comparisons of video generation performance.} TASTE-Rob generates videos with high-quality and accurate action, while other existing powerful general VDMs fail.}
   \label{fig:comp}
   \vspace{-5pt}
\end{figure*}

\subsection{Baselines and Comparisons.} We select four existing powerful I2V diffusion models - DynamiCrafter~\cite{xing2023dynamicrafteranimatingopendomainimages}, consistI2V~\cite{ren2024consisti2venhancingvisualconsistency}, Open-Sora Plan~\cite{pku_yuan_lab_and_tuzhan_ai_etc_2024_10948109} and CogVideoX~\cite{yang2024cogvideoxtexttovideodiffusionmodels}, as baselines, and conduct comparative experiments among these baselines and our method. We provide qualitative comparisons of video generation performance on \dataset{}-Test and wild environment in Figure \ref{fig:comp}, and quantitative comparisons on \dataset{}-Test in Table \ref{tab:vid_comp_baselines}, which demonstrates superior video quality and better generalization capability of our method. 
According to above experiments, all existing powerful general VDMs fail to achieve manipulation tasks well, making them unsuitable for generating HOI video demonstrations. Given that the other two evaluation aspects focus on measuring the fine-grained details of generated videos, we omit further comparisons with these baseline methods.


\subsection{Ablation Study}
\label{sec:ablation}

\begin{figure}[h]
  \centering
  \includegraphics[width=0.9\linewidth]{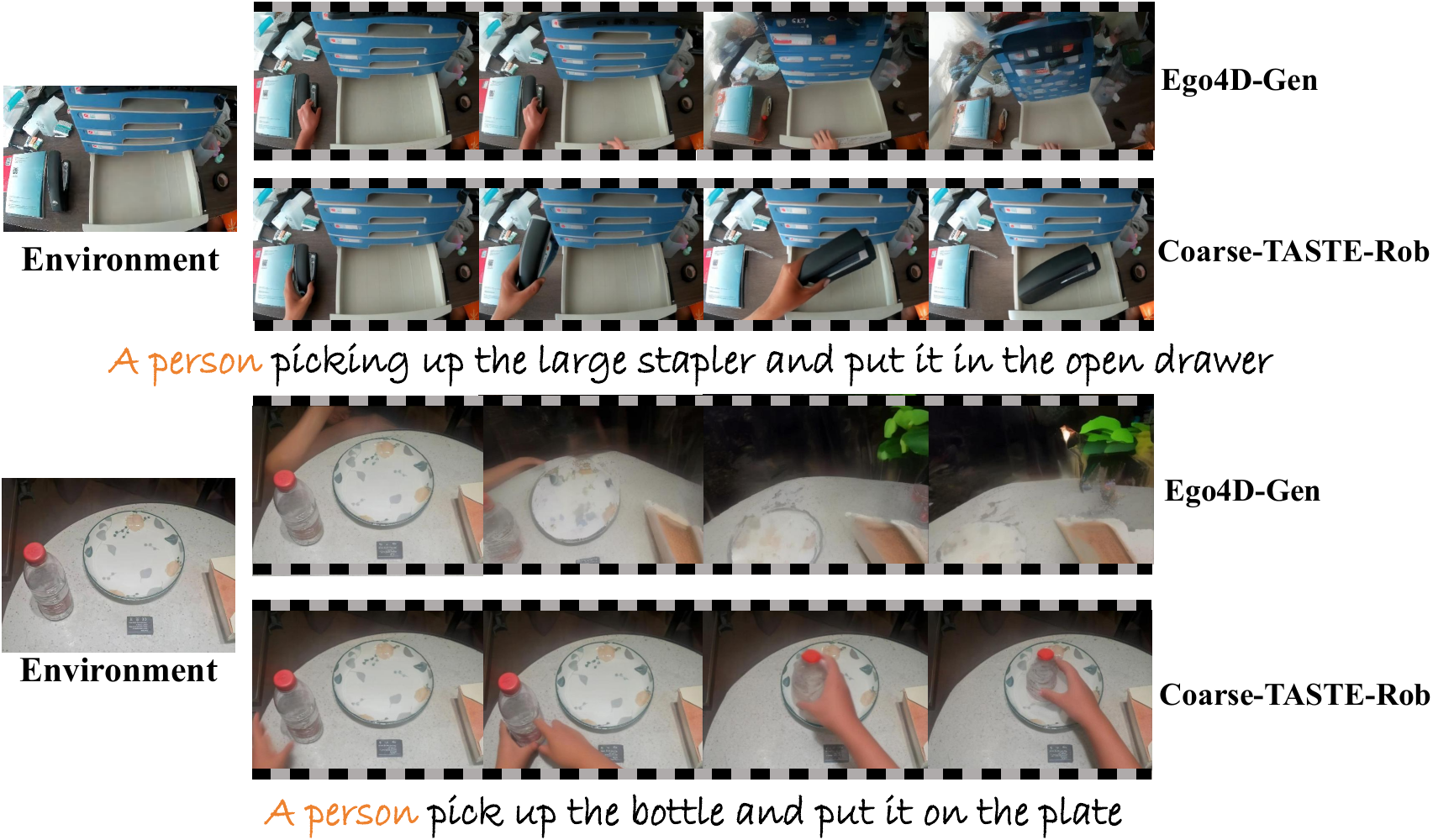}

   \vspace{-10pt}
   \caption{\textbf{Qualitative comparison between Ego4D-Gen and Coarse-TASTE-Rob.} Our \dataset{} dataset significantly improves the generalization of video generation models in generating HOI videos with accurate actions.} 
   \label{fig:Ego4D-Gen}
   \vspace{-15pt}
\end{figure}

\begin{table}[h]
 \centering
 \begin{tabular}{@{}cccc@{}}
   \toprule
   \multirow{2}{*}[-0.5ex]{\centering Method} & \multicolumn{3}{c}{Generation Quality} \\
   \cmidrule(l){2-4}
   & KVD$\downarrow$ & FVD$\downarrow$ & PIC$\uparrow$ \\
   \midrule
   Ego4D-Gen & 0.18 & 52.17 & 0.77 \\
   \textbf{Coarse-TASTE-Rob} & \textbf{0.04} & \textbf{10.85} & \textbf{0.88} \\
   \bottomrule
 \end{tabular}
 \vspace{-5pt}
 \caption{\textbf{Quantitative comparison between Ego4D-Gen and Coarse-TASTE-Rob.} All video generation metrics prove the better generation performance of Coarse-TASTE-Rob.}
 \label{tab:comp_dataset_gen}
  \vspace{-15pt}
\end{table}

\paragraph{\dataset{} Dataset.} We evaluate the performance improvements achieved by using our \dataset{} dataset. To be fair, we compare video generation performance between our coarse action planner (Coarse-\dataset{}) and Ego4D-Gen, a comparison baseline fine-tuned from DynamiCrafter~\cite{xing2023dynamicrafteranimatingopendomainimages} on 83,647 ego-centric HOI videos from Ego4D~\cite{grauman2022ego4dworld3000hours}. The only difference between them is the training datasets they use. As shown in Figure \ref{fig:Ego4D-Gen} and Table \ref{tab:comp_dataset_gen}, our coarse action planner achieves superior performance. 

\begin{table}[h]
 \centering
 \begin{tabular}{@{}ccc@{}}
   \toprule
   \centering Metric & Coarse-\dataset{} & \dataset{} \\
   \midrule
   KVD$\downarrow$ & 0.04 & \textbf{0.03} \\
   FVD$\downarrow$ & 10.85 & \textbf{9.43} \\
   PIC$\uparrow$ & 0.88 & \textbf{0.90} \\
   \midrule
   GPV$\downarrow$ & 0.28 & \textbf{0.24} \\
   GTCE$\downarrow$ & 67.8$\%$ & \textbf{9.7$\%$} \\
   HMDA$\downarrow$ & 26.4$^\circ$ & \textbf{11.3$^\circ$} \\
   \midrule 
   success rate$\uparrow$ & 84$\%$ & \textbf{96$\%$} \\
   \bottomrule
 \end{tabular}
 \vspace{-5pt}
 \caption{\textbf{Quantitative comparison between TASTE-Rob and Coarse-TASTE-Rob.} All metrics prove the better generation performance, grasping pose consistency and more precise robot manipulation of TASTE-Rob.}
 \vspace{-15pt}
 \label{tab:comp_pipeline}

\end{table}

\begin{figure}[h]
  \centering
  \includegraphics[width=0.9\linewidth]{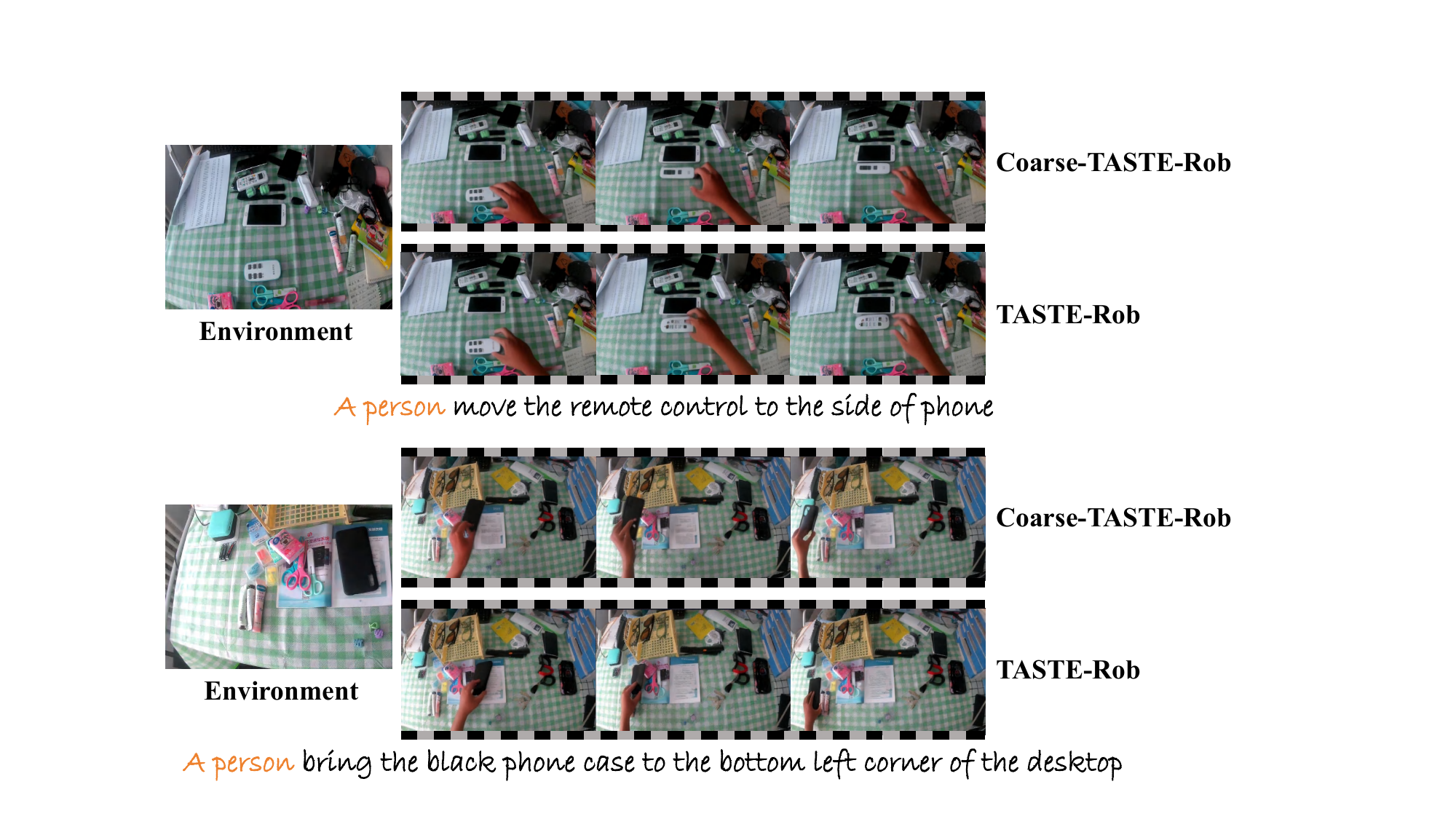}

   \vspace{-10pt}
    \caption{\textbf{Qualitative comparison between TASTE-Rob and Coarse-TASTE-Rob.} With our proposed pose-refinement pipeline, we can generate HOI videos with more accurate and feasible grasping poses.}
   \label{fig:pose-comp}
   \vspace{-15pt}
\end{figure}

\begin{figure}[h]
  \centering
  \includegraphics[width=0.9\linewidth]{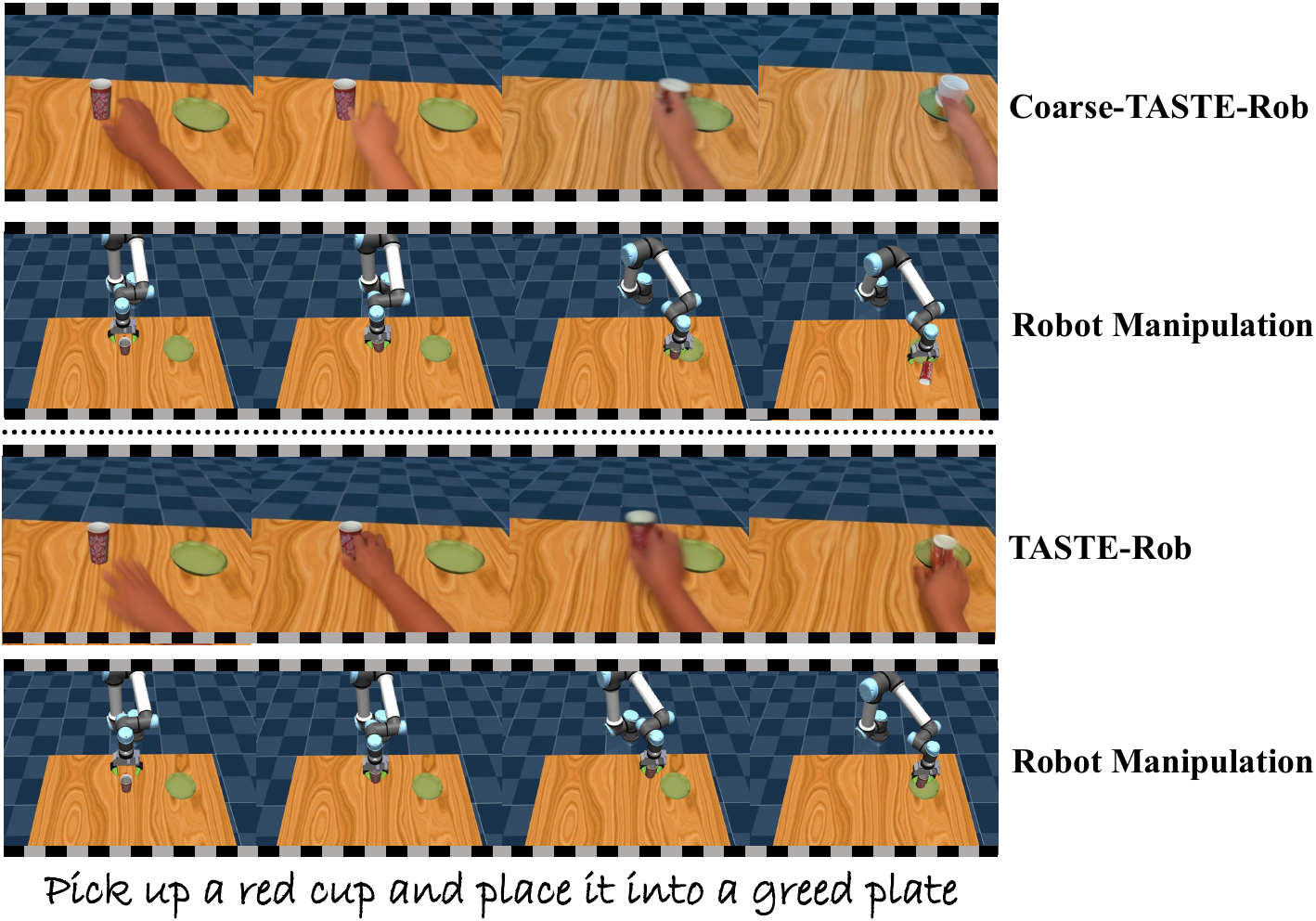}
    \vspace{-10pt}
   \caption{\textbf{Comparison of Robot Manipulation Performance.} TASTE-Rob generates higher-quality HOI videos, leading to more precise robot manipulation.}
   \label{fig:RM_case}
\end{figure}

\paragraph{Pose-Refinement Pipeline.} We compare Coarse-\dataset{} and \dataset{} across all three evaluation aspects, as shown in Table \ref{tab:comp_pipeline}. \textbf{At a coarse level}, both the video generation metrics in Table \ref{tab:comp_pipeline} and the visual comparisons in Figure \ref{fig:pose-comp} demonstrate that our proposed pose-refinement pipeline achieves better HOI video generation. \textbf{At a fine level}, both the grasping pose
consistency metrics in Table \ref{tab:comp_pipeline} and Figure \ref{fig:pose-comp} further validate that our proposed pose-refinement pipeline effectively addresses the pose inconsistency issue and enables more precise robot manipulation. Besides, as shown in Figure \ref{fig:RM_case} and success rate metrics in Table \ref{tab:comp_pipeline}, our proposed pose-refinement pipeline helps the policy model achieve more accurate target placement, leading to improved success rates. In addition, we only do this ablation on our dataset, because videos generated by fine-tuning on existing HOI datasets (e.g., Ego4D~\cite{grauman2022ego4dworld3000hours}) are of low quality so that it’s difficult to extract reasonable poses for further refinement.

\paragraph{Denoising Steps of Pose Refinement.} We also discuss the selection of $N_{rv}$ during pose refinement. When $N_{rv}=0$ (no refinement), hand pose sequences exhibit inconsistency issues, while $N_{rv}=50$ (generating pose sequences from Gaussian noise) results in limited spatial awareness. As $N_{rv}$ increases, we observe improved pose consistency but degraded spatial awareness. As shown in Table~\ref{tab:comp_dataset}, when $N_{rv}$ increases, video generation quality first improves and then deteriorates. Based on this trade-off, 
we finally select the value of $N_{rv}$ as 10 
(see more details in Appendix \ref{appendix:result}).

\section{Discussion and Conclusion}

\paragraph{Summary.} In this work, we collect a diverse, large-scale ego-centric HOI dataset and develop a three-stage pose-refinement pipeline to generate high-quality task-oriented HOI videos. We successfully enhance HOI video generation capabilities, thereby improving robot manipulation generalization to novel scenes through imitation learning.

\paragraph{Limitations.} Our approach achieves accurate manipulation on unseen scenes and enhances the grasping poses in generated videos. However, the generation quality degrades when objects undergo significant transformations, such as rotation and opening. In our future work, we plan to address this limitation. 
\clearpage
\section{Acknowledgments}
The work was supported in part by the Basic Research Project
No. HZQB-KCZYZ-2021067 of Hetao Shenzhen-HK S\&T Cooperation Zone, Guangdong Provincial Outstanding Youth Fund(No.
2023B1515020055), NSFC-61931024, and Shenzhen Science and Technology Program No. JCYJ20220530143604010. It is also partly supported by the National Key R\&D Program of China with grant
No. 2018YFB1800800, by Shenzhen Outstanding Talents Training
Fund 202002, by Guangdong Research Projects No. 2017ZT07X152
and No. 2019CX01X104,by Key Area R\&D Program of Guangdong Province (Grant No. 2018B030338001), by the Guangdong Provincial Key Laboratory of Future Networks of Intelligence (Grant No.
2022B1212010001), and by Shenzhen Key Laboratory of Big Data
and Artificial Intelligence (Grant No. ZDSYS201707251409055).

{
    \small
    \bibliographystyle{ieeenat_fullname}
    \bibliography{main}

\begin{thebibliography}{65}
\providecommand{\natexlab}[1]{#1}
\providecommand{\url}[1]{\texttt{#1}}
\expandafter\ifx\csname urlstyle\endcsname\relax
  \providecommand{\doi}[1]{doi: #1}\else
  \providecommand{\doi}{doi: \begingroup \urlstyle{rm}\Url}\fi

\bibitem[Bahl et~al.(2022)Bahl, Gupta, and Pathak]{bahl2022human}
Shikhar Bahl, Abhinav Gupta, and Deepak Pathak.
\newblock Human-to-robot imitation in the wild.
\newblock 2022.

\bibitem[Bharadhwaj et~al.(2023{\natexlab{a}})Bharadhwaj, Gupta, Kumar, and Tulsiani]{bharadhwaj2023generalizablezeroshotmanipulationtranslating}
Homanga Bharadhwaj, Abhinav Gupta, Vikash Kumar, and Shubham Tulsiani.
\newblock Towards generalizable zero-shot manipulation via translating human interaction plans, 2023{\natexlab{a}}.

\bibitem[Bharadhwaj et~al.(2023{\natexlab{b}})Bharadhwaj, Gupta, and Tulsiani]{BharadhwajVisual}
Homanga Bharadhwaj, Abhinav Gupta, and Shubham Tulsiani.
\newblock Visual affordance prediction for guiding robot exploration.
\newblock \emph{IEEE International Conference on Robotics and Automation (ICRA)}, 2023{\natexlab{b}}.

\bibitem[Bharadhwaj et~al.(2023{\natexlab{c}})Bharadhwaj, Vakil, Sharma, Gupta, Tulsiani, and Kumar]{bharadhwaj2023roboagentgeneralizationefficiencyrobot}
Homanga Bharadhwaj, Jay Vakil, Mohit Sharma, Abhinav Gupta, Shubham Tulsiani, and Vikash Kumar.
\newblock Roboagent: Generalization and efficiency in robot manipulation via semantic augmentations and action chunking, 2023{\natexlab{c}}.

\bibitem[Bharadhwaj et~al.(2024{\natexlab{a}})Bharadhwaj, Dwibedi, Gupta, Tulsiani, Doersch, Xiao, Shah, Xia, Sadigh, and Kirmani]{bharadhwaj2024gen2acthumanvideogeneration}
Homanga Bharadhwaj, Debidatta Dwibedi, Abhinav Gupta, Shubham Tulsiani, Carl Doersch, Ted Xiao, Dhruv Shah, Fei Xia, Dorsa Sadigh, and Sean Kirmani.
\newblock Gen2act: Human video generation in novel scenarios enables generalizable robot manipulation, 2024{\natexlab{a}}.

\bibitem[Bharadhwaj et~al.(2024{\natexlab{b}})Bharadhwaj, Mottaghi, Gupta, and Tulsiani]{bharadhwaj2024track2act}
Homanga Bharadhwaj, Roozbeh Mottaghi, Abhinav Gupta, and Shubham Tulsiani.
\newblock Track2act: Predicting point tracks from internet videos enables generalizable robot manipulation.
\newblock In \emph{European Conference on Computer Vision (ECCV)}, 2024{\natexlab{b}}.

\bibitem[Black et~al.(2023)Black, Nakamoto, Atreya, Walke, Finn, Kumar, and Levine]{black2023zeroshotroboticmanipulationpretrained}
Kevin Black, Mitsuhiko Nakamoto, Pranav Atreya, Homer Walke, Chelsea Finn, Aviral Kumar, and Sergey Levine.
\newblock Zero-shot robotic manipulation with pretrained image-editing diffusion models, 2023.

\bibitem[Brohan et~al.(2022)Brohan, Brown, Carbajal, and \etal]{rt12022arxiv}
Anthony Brohan, Noah Brown, Justice Carbajal, and \etal.
\newblock Rt-1: Robotics transformer for real-world control at scale.
\newblock In \emph{arXiv preprint arXiv:2212.06817}, 2022.

\bibitem[Chane-Sane et~al.(2023)Chane-Sane, Schmid, and Laptev]{chanesane2023vip}
Elliot Chane-Sane, Cordelia Schmid, and Ivan Laptev.
\newblock Learning video-conditioned policies for unseen manipulation tasks.
\newblock In \emph{ICRA}, 2023.

\bibitem[Chi et~al.(2023)Chi, Feng, Du, Xu, Cousineau, Burchfiel, and Song]{chi2023diffusionpolicy}
Cheng Chi, Siyuan Feng, Yilun Du, Zhenjia Xu, Eric Cousineau, Benjamin Burchfiel, and Shuran Song.
\newblock Diffusion policy: Visuomotor policy learning via action diffusion.
\newblock In \emph{Proceedings of Robotics: Science and Systems (RSS)}, 2023.

\bibitem[Damen et~al.(2022)Damen, Doughty, Farinella, Furnari, Ma, Kazakos, Moltisanti, Munro, Perrett, Price, and Wray]{Damen2022RESCALING}
Dima Damen, Hazel Doughty, Giovanni~Maria Farinella, Antonino Furnari, Jian Ma, Evangelos Kazakos, Davide Moltisanti, Jonathan Munro, Toby Perrett, Will Price, and Michael Wray.
\newblock Rescaling egocentric vision: Collection, pipeline and challenges for epic-kitchens-100.
\newblock \emph{International Journal of Computer Vision (IJCV)}, 130:\penalty0 33–55, 2022.

\bibitem[Du et~al.(2023)Du, Yang, Dai, Dai, Nachum, Tenenbaum, Schuurmans, and Abbeel]{du2023learninguniversalpoliciestextguided}
Yilun Du, Mengjiao Yang, Bo Dai, Hanjun Dai, Ofir Nachum, Joshua~B. Tenenbaum, Dale Schuurmans, and Pieter Abbeel.
\newblock Learning universal policies via text-guided video generation, 2023.

\bibitem[Fathi et~al.(2012)Fathi, Hodgins, and Rehg]{6247805}
Alircza Fathi, Jessica~K. Hodgins, and James~M. Rehg.
\newblock Social interactions: A first-person perspective.
\newblock In \emph{2012 IEEE Conference on Computer Vision and Pattern Recognition}, pages 1226--1233, 2012.

\bibitem[Feix et~al.(2016)Feix, Romero, Schmiedmayer, Dollar, and Kragic]{7243327}
Thomas Feix, Javier Romero, Heinz-Bodo Schmiedmayer, Aaron~M. Dollar, and Danica Kragic.
\newblock The grasp taxonomy of human grasp types.
\newblock \emph{IEEE Transactions on Human-Machine Systems}, 46\penalty0 (1):\penalty0 66--77, 2016.

\bibitem[Florence et~al.(2021)Florence, Lynch, Zeng, Ramirez, Wahid, Downs, Wong, Lee, Mordatch, and Tompson]{florence2021implicit}
Pete Florence, Corey Lynch, Andy Zeng, Oscar Ramirez, Ayzaan Wahid, Laura Downs, Adrian Wong, Johnny Lee, Igor Mordatch, and Jonathan Tompson.
\newblock Implicit behavioral cloning.
\newblock \emph{Conference on Robot Learning (CoRL)}, 2021.

\bibitem[Grauman et~al.(2022)Grauman, Westbury, and \etal]{grauman2022ego4dworld3000hours}
Kristen Grauman, Andrew Westbury, and \etal.
\newblock Ego4d: Around the world in 3,000 hours of egocentric video, 2022.

\bibitem[Ho et~al.(2020)Ho, Jain, and Abbeel]{ho2020denoisingdiffusionprobabilisticmodels}
Jonathan Ho, Ajay Jain, and Pieter Abbeel.
\newblock Denoising diffusion probabilistic models, 2020.

\bibitem[Ho et~al.(2022)Ho, Salimans, Gritsenko, Chan, Norouzi, and Fleet]{ho2022videodiffusionmodels}
Jonathan Ho, Tim Salimans, Alexey Gritsenko, William Chan, Mohammad Norouzi, and David~J. Fleet.
\newblock Video diffusion models, 2022.

\bibitem[Hu et~al.(2023)Hu, Yin, Ma, Chen, Fernando, Asano, Gavves, Mettes, Ommer, and Snoek]{hu2023motionflowmatchinghuman}
Vincent~Tao Hu, Wenzhe Yin, Pingchuan Ma, Yunlu Chen, Basura Fernando, Yuki~M Asano, Efstratios Gavves, Pascal Mettes, Bjorn Ommer, and Cees G.~M. Snoek.
\newblock Motion flow matching for human motion synthesis and editing, 2023.

\bibitem[Jain et~al.(2024)Jain, Attarian, Joshi, Wahid, Driess, Vuong, Sanketi, Sermanet, Welker, Chan, Gilitschenski, Bisk, and Dwibedi]{jain2024vid2robotendtoendvideoconditionedpolicy}
Vidhi Jain, Maria Attarian, Nikhil~J Joshi, Ayzaan Wahid, Danny Driess, Quan Vuong, Pannag~R Sanketi, Pierre Sermanet, Stefan Welker, Christine Chan, Igor Gilitschenski, Yonatan Bisk, and Debidatta Dwibedi.
\newblock Vid2robot: End-to-end video-conditioned policy learning with cross-attention transformers, 2024.

\bibitem[Jang et~al.(2021)Jang, Irpan, Khansari, Kappler, Ebert, Lynch, Levine, and Finn]{jang2021bc}
Eric Jang, Alex Irpan, Mohi Khansari, Daniel Kappler, Frederik Ebert, Corey Lynch, Sergey Levine, and Chelsea Finn.
\newblock {BC}-z: Zero-shot task generalization with robotic imitation learning.
\newblock In \emph{5th Annual Conference on Robot Learning}, 2021.

\bibitem[Jonnavittula et~al.(2024)Jonnavittula, Parekh, and Losey]{jonnavittula2024viewvisualimitationlearning}
Ananth Jonnavittula, Sagar Parekh, and Dylan~P. Losey.
\newblock View: Visual imitation learning with waypoints, 2024.

\bibitem[Karamcheti et~al.(2023)Karamcheti, Nair, Chen, Kollar, Finn, Sadigh, and Liang]{karamcheti2023voltron}
Siddharth Karamcheti, Suraj Nair, Annie~S. Chen, Thomas Kollar, Chelsea Finn, Dorsa Sadigh, and Percy Liang.
\newblock Language-driven representation learning for robotics.
\newblock In \emph{Robotics: Science and Systems (RSS)}, 2023.

\bibitem[Khazatsky et~al.(2024)Khazatsky, Pertsch, and \etal]{khazatsky2024droidlargescaleinthewildrobot}
Alexander Khazatsky, Karl Pertsch, and \etal.
\newblock Droid: A large-scale in-the-wild robot manipulation dataset, 2024.

\bibitem[Ko et~al.(2023)Ko, Mao, Du, Sun, and Tenenbaum]{ko2023learningactactionlessvideos}
Po-Chen Ko, Jiayuan Mao, Yilun Du, Shao-Hua Sun, and Joshua~B. Tenenbaum.
\newblock Learning to act from actionless videos through dense correspondences, 2023.

\bibitem[Lab and etc.(2024)]{pku_yuan_lab_and_tuzhan_ai_etc_2024_10948109}
PKU-Yuan Lab and Tuzhan~AI etc.
\newblock Open-sora-plan, 2024.

\bibitem[Lee et~al.(2012)Lee, Ghosh, and Grauman]{6247820}
Yong~Jae Lee, Joydeep Ghosh, and Kristen Grauman.
\newblock Discovering important people and objects for egocentric video summarization.
\newblock In \emph{2012 IEEE Conference on Computer Vision and Pattern Recognition}, pages 1346--1353, 2012.

\bibitem[Li et~al.(2020)Li, Liu, and Rehg]{li2020eyebeholdergazeactions}
Yin Li, Miao Liu, and James~M. Rehg.
\newblock In the eye of the beholder: Gaze and actions in first person video, 2020.

\bibitem[Liu et~al.(2022)Liu, Liu, Jiang, Lyu, Wan, Shen, Liang, Fu, Wang, and Yi]{Liu_2022_CVPR}
Yunze Liu, Yun Liu, Che Jiang, Kangbo Lyu, Weikang Wan, Hao Shen, Boqiang Liang, Zhoujie Fu, He Wang, and Li Yi.
\newblock Hoi4d: A 4d egocentric dataset for category-level human-object interaction.
\newblock In \emph{Proceedings of the IEEE/CVF Conference on Computer Vision and Pattern Recognition (CVPR)}, pages 21013--21022, 2022.

\bibitem[Lu and Grauman(2013)]{6619194}
Zheng Lu and Kristen Grauman.
\newblock Story-driven summarization for egocentric video.
\newblock In \emph{2013 IEEE Conference on Computer Vision and Pattern Recognition}, pages 2714--2721, 2013.

\bibitem[Lugaresi et~al.(2019)Lugaresi, Tang, Nash, McClanahan, Uboweja, Hays, Zhang, Chang, Yong, Lee, Chang, Hua, Georg, and Grundmann]{lugaresi2019mediapipeframeworkbuildingperception}
Camillo Lugaresi, Jiuqiang Tang, Hadon Nash, Chris McClanahan, Esha Uboweja, Michael Hays, Fan Zhang, Chuo-Ling Chang, Ming~Guang Yong, Juhyun Lee, Wan-Teh Chang, Wei Hua, Manfred Georg, and Matthias Grundmann.
\newblock Mediapipe: A framework for building perception pipelines, 2019.

\bibitem[Ma et~al.(2023)Ma, Liang, Som, Kumar, Zhang, Bastani, and Jayaraman]{ma2023livlanguageimagerepresentationsrewards}
Yecheng~Jason Ma, William Liang, Vaidehi Som, Vikash Kumar, Amy Zhang, Osbert Bastani, and Dinesh Jayaraman.
\newblock Liv: Language-image representations and rewards for robotic control, 2023.

\bibitem[Mou et~al.(2023)Mou, Wang, Xie, Wu, Zhang, Qi, Shan, and Qie]{mou2023t2iadapterlearningadaptersdig}
Chong Mou, Xintao Wang, Liangbin Xie, Yanze Wu, Jian Zhang, Zhongang Qi, Ying Shan, and Xiaohu Qie.
\newblock T2i-adapter: Learning adapters to dig out more controllable ability for text-to-image diffusion models, 2023.

\bibitem[Pavlakos et~al.(2024)Pavlakos, Shan, Radosavovic, Kanazawa, Fouhey, and Malik]{pavlakos2024reconstructing}
Georgios Pavlakos, Dandan Shan, Ilija Radosavovic, Angjoo Kanazawa, David Fouhey, and Jitendra Malik.
\newblock Reconstructing hands in 3{D} with transformers.
\newblock In \emph{CVPR}, 2024.

\bibitem[Pirsiavash and Ramanan(2012)]{6248010}
Hamed Pirsiavash and Deva Ramanan.
\newblock Detecting activities of daily living in first-person camera views.
\newblock In \emph{2012 IEEE Conference on Computer Vision and Pattern Recognition}, pages 2847--2854, 2012.

\bibitem[Raab et~al.(2024)Raab, Leibovitch, Tevet, Arar, Bermano, and Cohen-Or]{raab2024single}
Sigal Raab, Inbal Leibovitch, Guy Tevet, Moab Arar, Amit~H Bermano, and Daniel Cohen-Or.
\newblock Single motion diffusion.
\newblock In \emph{The Twelfth International Conference on Learning Representations (ICLR)}, 2024.

\bibitem[Radford et~al.(2021)Radford, Kim, Hallacy, Ramesh, Goh, Agarwal, Sastry, Askell, Mishkin, Clark, Krueger, and Sutskever]{radford2021learningtransferablevisualmodels}
Alec Radford, Jong~Wook Kim, Chris Hallacy, Aditya Ramesh, Gabriel Goh, Sandhini Agarwal, Girish Sastry, Amanda Askell, Pamela Mishkin, Jack Clark, Gretchen Krueger, and Ilya Sutskever.
\newblock Learning transferable visual models from natural language supervision, 2021.

\bibitem[Ren et~al.(2024)Ren, Yang, Zhang, Wei, Du, Huang, and Chen]{ren2024consisti2venhancingvisualconsistency}
Weiming Ren, Huan Yang, Ge Zhang, Cong Wei, Xinrun Du, Wenhao Huang, and Wenhu Chen.
\newblock Consisti2v: Enhancing visual consistency for image-to-video generation, 2024.

\bibitem[Romero et~al.(2017)Romero, Tzionas, and Black]{Romero_2017}
Javier Romero, Dimitrios Tzionas, and Michael~J. Black.
\newblock Embodied hands: modeling and capturing hands and bodies together.
\newblock \emph{ACM Transactions on Graphics}, 36\penalty0 (6):\penalty0 1–17, 2017.

\bibitem[Shan et~al.(2020)Shan, Geng, Shu, and Fouhey]{Shan20}
Dandan Shan, Jiaqi Geng, Michelle Shu, and David Fouhey.
\newblock Understanding human hands in contact at internet scale.
\newblock In \emph{CVPR}, 2020.

\bibitem[Shi et~al.(2023)Shi, Xiong, Lin, and Jung]{shi2023instantboothpersonalizedtexttoimagegeneration}
Jing Shi, Wei Xiong, Zhe Lin, and Hyun~Joon Jung.
\newblock Instantbooth: Personalized text-to-image generation without test-time finetuning, 2023.

\bibitem[Shridhar et~al.(2022)Shridhar, Manuelli, and Fox]{shridhar2022peract}
Mohit Shridhar, Lucas Manuelli, and Dieter Fox.
\newblock Perceiver-actor: A multi-task transformer for robotic manipulation.
\newblock In \emph{Proceedings of the 6th Conference on Robot Learning (CoRL)}, 2022.

\bibitem[Sigurdsson et~al.(2018)Sigurdsson, Gupta, Schmid, Farhadi, and Alahari]{sigurdsson2018charadesegolargescaledatasetpaired}
Gunnar~A. Sigurdsson, Abhinav Gupta, Cordelia Schmid, Ali Farhadi, and Karteek Alahari.
\newblock Charades-ego: A large-scale dataset of paired third and first person videos, 2018.

\bibitem[Sohl-Dickstein et~al.(2015)Sohl-Dickstein, Weiss, Maheswaranathan, and Ganguli]{sohldickstein2015deepunsupervisedlearningusing}
Jascha Sohl-Dickstein, Eric~A. Weiss, Niru Maheswaranathan, and Surya Ganguli.
\newblock Deep unsupervised learning using nonequilibrium thermodynamics, 2015.

\bibitem[Tevet et~al.(2023)Tevet, Raab, Gordon, Shafir, Cohen-or, and Bermano]{tevet2023human}
Guy Tevet, Sigal Raab, Brian Gordon, Yoni Shafir, Daniel Cohen-or, and Amit~Haim Bermano.
\newblock Human motion diffusion model.
\newblock In \emph{The Eleventh International Conference on Learning Representations}, 2023.

\bibitem[Unterthiner et~al.(2019)Unterthiner, van Steenkiste, Kurach, Marinier, Michalski, and Gelly]{Unterthiner2019FVDAN}
Thomas Unterthiner, Sjoerd van Steenkiste, Karol Kurach, Rapha{\"e}l Marinier, Marcin Michalski, and Sylvain Gelly.
\newblock Fvd: A new metric for video generation.
\newblock In \emph{DGS@ICLR}, 2019.

\bibitem[Wang et~al.(2024{\natexlab{a}})Wang, Sridhar, Feng, der Merwe, Fishman, Fazeli, and Park]{wang2024thisthatlanguagegesturecontrolledvideo}
Boyang Wang, Nikhil Sridhar, Chao Feng, Mark~Van der Merwe, Adam Fishman, Nima Fazeli, and Jeong~Joon Park.
\newblock This\&that: Language-gesture controlled video generation for robot planning, 2024{\natexlab{a}}.

\bibitem[Wang et~al.(2023{\natexlab{a}})Wang, Fan, Sun, Zhang, Fei-Fei, Xu, Zhu, and Anandkumar]{wang2023mimicplaylonghorizonimitationlearning}
Chen Wang, Linxi Fan, Jiankai Sun, Ruohan Zhang, Li Fei-Fei, Danfei Xu, Yuke Zhu, and Anima Anandkumar.
\newblock Mimicplay: Long-horizon imitation learning by watching human play, 2023{\natexlab{a}}.

\bibitem[Wang et~al.(2023{\natexlab{b}})Wang, Yuan, Zhang, Chen, Wang, Zhang, Shen, Zhao, and Zhou]{wang2023videocomposercompositionalvideosynthesis}
Xiang Wang, Hangjie Yuan, Shiwei Zhang, Dayou Chen, Jiuniu Wang, Yingya Zhang, Yujun Shen, Deli Zhao, and Jingren Zhou.
\newblock Videocomposer: Compositional video synthesis with motion controllability, 2023{\natexlab{b}}.

\bibitem[Wang et~al.(2023{\natexlab{c}})Wang, Leng, Li, Wu, and Liang]{10376967}
Yin Wang, Zhiying Leng, Frederick W.~B. Li, Shun-Cheng Wu, and Xiaohui Liang.
\newblock Fg-t2m: Fine-grained text-driven human motion generation via diffusion model.
\newblock In \emph{2023 IEEE/CVF International Conference on Computer Vision (ICCV)}, pages 21978--21987, 2023{\natexlab{c}}.

\bibitem[Wang et~al.(2024{\natexlab{b}})Wang, Zheng, Nie, Xu, Wang, Ye, Li, Zhang, Cheng, Dong, Cai, Lin, Zheng, and Liang]{wang2024robotsonenewstandard}
Zhiqiang Wang, Hao Zheng, Yunshuang Nie, Wenjun Xu, Qingwei Wang, Hua Ye, Zhe Li, Kaidong Zhang, Xuewen Cheng, Wanxi Dong, Chang Cai, Liang Lin, Feng Zheng, and Xiaodan Liang.
\newblock All robots in one: A new standard and unified dataset for versatile, general-purpose embodied agents, 2024{\natexlab{b}}.

\bibitem[Wen et~al.(2024)Wen, Lin, So, Chen, Dou, Gao, and Abbeel]{wen2024anypointtrajectorymodelingpolicy}
Chuan Wen, Xingyu Lin, John So, Kai Chen, Qi Dou, Yang Gao, and Pieter Abbeel.
\newblock Any-point trajectory modeling for policy learning, 2024.

\bibitem[Wu et~al.(2024)Wu, Jing, Cheang, Chen, Xu, Li, Liu, Li, and Kong]{wu2023unleashing}
Hongtao Wu, Ya Jing, Chilam Cheang, Guangzeng Chen, Jiafeng Xu, Xinghang Li, Minghuan Liu, Hang Li, and Tao Kong.
\newblock Unleashing large-scale video generative pre-training for visual robot manipulation.
\newblock In \emph{International Conference on Learning Representations}, 2024.

\bibitem[Xing et~al.(2023)Xing, Xia, Zhang, Chen, Yu, Liu, Wang, Wong, and Shan]{xing2023dynamicrafteranimatingopendomainimages}
Jinbo Xing, Menghan Xia, Yong Zhang, Haoxin Chen, Wangbo Yu, Hanyuan Liu, Xintao Wang, Tien-Tsin Wong, and Ying Shan.
\newblock Dynamicrafter: Animating open-domain images with video diffusion priors, 2023.

\bibitem[Xing et~al.(2024)Xing, Liu, Xia, Zhang, Wang, Shan, and Wong]{xing2024tooncrafter}
Jinbo Xing, Hanyuan Liu, Menghan Xia, Yong Zhang, Xintao Wang, Ying Shan, and Tien-Tsin Wong.
\newblock Tooncrafter: Generative cartoon interpolation.
\newblock \emph{ACM Transactions on Graphics (TOG)}, 43\penalty0 (6):\penalty0 1--11, 2024.

\bibitem[Xiong et~al.(2021)Xiong, Li, Chen, Bharadhwaj, Sinha, and Garg]{xiong2021learningwatchingphysicalimitation}
Haoyu Xiong, Quanzhou Li, Yun-Chun Chen, Homanga Bharadhwaj, Samarth Sinha, and Animesh Garg.
\newblock Learning by watching: Physical imitation of manipulation skills from human videos, 2021.

\bibitem[Xu et~al.(2023)Xu, Xu, Chi, Veloso, and Song]{xu2023xskill}
Mengda Xu, Zhenjia Xu, Cheng Chi, Manuela Veloso, and Shuran Song.
\newblock {XS}kill: Cross embodiment skill discovery.
\newblock In \emph{7th Annual Conference on Robot Learning}, 2023.

\bibitem[Xu et~al.(2024)Xu, Xu, Xu, Chi, Wetzstein, Veloso, and Song]{xu2024flow}
Mengda Xu, Zhenjia Xu, Yinghao Xu, Cheng Chi, Gordon Wetzstein, Manuela Veloso, and Shuran Song.
\newblock Flow as the cross-domain manipulation interface.
\newblock In \emph{8th Annual Conference on Robot Learning}, 2024.

\bibitem[Yang et~al.(2023)Yang, Su, and Wen]{yang2023synthesizinglongtermhumanmotions}
Zhao Yang, Bing Su, and Ji-Rong Wen.
\newblock Synthesizing long-term human motions with diffusion models via coherent sampling, 2023.

\bibitem[Yang et~al.(2024)Yang, Teng, Zheng, Ding, Huang, Xu, Yang, Hong, Zhang, Feng, Yin, Gu, Zhang, Wang, Cheng, Liu, Xu, Dong, and Tang]{yang2024cogvideoxtexttovideodiffusionmodels}
Zhuoyi Yang, Jiayan Teng, Wendi Zheng, Ming Ding, Shiyu Huang, Jiazheng Xu, Yuanming Yang, Wenyi Hong, Xiaohan Zhang, Guanyu Feng, Da Yin, Xiaotao Gu, Yuxuan Zhang, Weihan Wang, Yean Cheng, Ting Liu, Bin Xu, Yuxiao Dong, and Jie Tang.
\newblock Cogvideox: Text-to-video diffusion models with an expert transformer, 2024.

\bibitem[Ye et~al.(2023)Ye, Zhang, Liu, Han, and Yang]{ye2023ipadaptertextcompatibleimage}
Hu Ye, Jun Zhang, Sibo Liu, Xiao Han, and Wei Yang.
\newblock Ip-adapter: Text compatible image prompt adapter for text-to-image diffusion models, 2023.

\bibitem[Zhang et~al.(2023{\natexlab{a}})Zhang, Rao, and Agrawala]{zhang2023adding}
Lvmin Zhang, Anyi Rao, and Maneesh Agrawala.
\newblock Adding conditional control to text-to-image diffusion models, 2023{\natexlab{a}}.

\bibitem[Zhang et~al.(2023{\natexlab{b}})Zhang, Wang, Zhang, Zhao, Yuan, Qin, Wang, Zhao, and Zhou]{zhang2023i2vgenxlhighqualityimagetovideosynthesis}
Shiwei Zhang, Jiayu Wang, Yingya Zhang, Kang Zhao, Hangjie Yuan, Zhiwu Qin, Xiang Wang, Deli Zhao, and Jingren Zhou.
\newblock I2vgen-xl: High-quality image-to-video synthesis via cascaded diffusion models, 2023{\natexlab{b}}.

\bibitem[Zhao et~al.(2023)Zhao, Kumar, Levine, and Finn]{zhao2023learningfinegrainedbimanualmanipulation}
Tony~Z. Zhao, Vikash Kumar, Sergey Levine, and Chelsea Finn.
\newblock Learning fine-grained bimanual manipulation with low-cost hardware, 2023.

\bibitem[Zhu et~al.(2024)Zhu, Lim, Stone, and Zhu]{zhu2024visionbasedmanipulationsinglehuman}
Yifeng Zhu, Arisrei Lim, Peter Stone, and Yuke Zhu.
\newblock Vision-based manipulation from single human video with open-world object graphs, 2024.

\end{thebibliography}
}

\clearpage
\setcounter{page}{1}
\maketitlesupplementary

\section{Details of \dataset{} dataset}
\label{appendix:dataset}

\subsection{Comparisons between Ego4D and \dataset{}}
\begin{figure}[h]
  \centering
  \begin{subfigure}{1.0\linewidth}
    \includegraphics[width=1.0\linewidth]{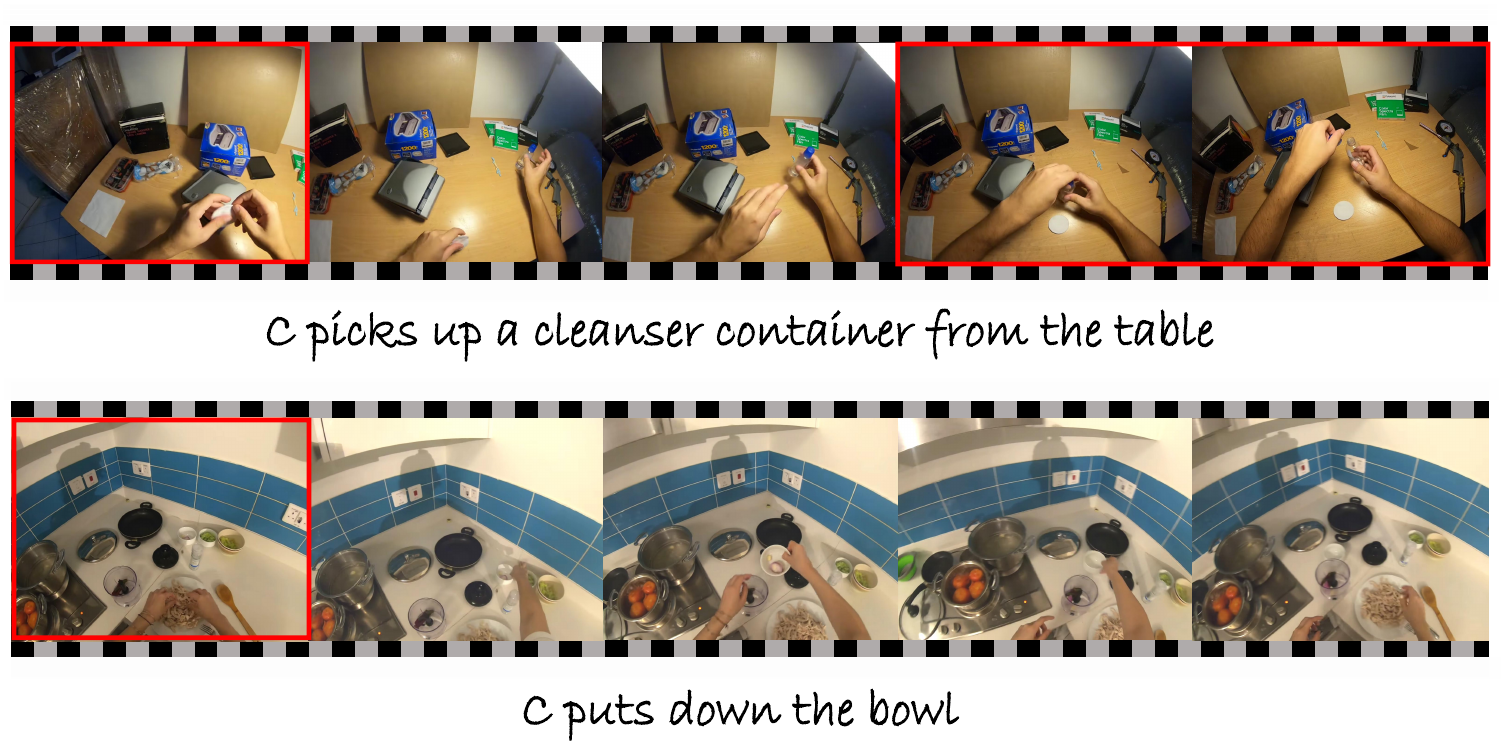}
    \caption{Examples of HOI videos in Ego4D~\cite{grauman2022ego4dworld3000hours}.}
    \label{fig:ex_ego4d}
  \end{subfigure}
  \hfill
  \begin{subfigure}{1.0\linewidth}
    \includegraphics[width=1.0\linewidth]{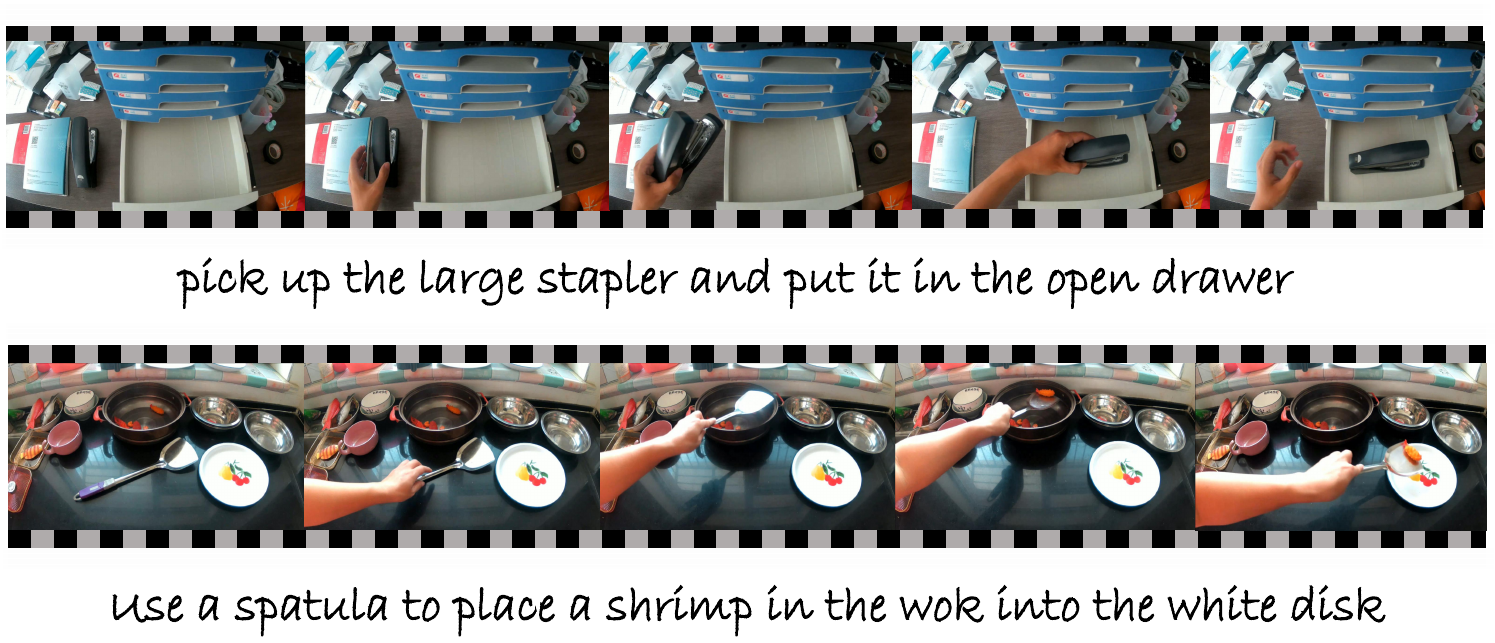}
    \caption{Examples of HOI videos in \dataset{}.}
    \label{fig:ex_ours}
  \end{subfigure}
  \vspace{-10pt}
  \caption{\textbf{Comparison between Ego4D~\cite{grauman2022ego4dworld3000hours} and \dataset{}}. Ego4D videos lack fixed camera perspective and precise action-language alignment, as shown in the red boxes, while \dataset{} solve these limitations.}
  \label{fig:example}
  \vspace{-15pt}
\end{figure}

Ego4D~\cite{grauman2022ego4dworld3000hours} differs from \dataset{} in two key aspects: the camera perspective and the recording methodology. \textbf{Camera Perspective}: Ego4D utilized seven distinct head-mounted cameras during data collection, resulting in diverse camera perspectives throughout the dataset. In contrast, a consistent camera perspective is essential for our task. As demonstrated in Figure~\ref{fig:example}, videos in Ego4D show perspective variations in the positioning of static objects, as evidenced by the blue box in the first case. \textbf{Recording Methodology}: In Ego4D's data collection process, participants were provided with cameras to record their daily activities. These recordings typically exceeded eight minutes in duration, and final short clips are extracted from them. Consequently, these clips often contain actions from preceding or subsequent activities, creating misalignment with the provided language instructions. This misalignment is exemplified in Figure~\ref{fig:example} (second case, highlighted by red bounding boxes): the participant is mixing food on a plate, which contradicts the instruction ``puts down the bowl''. 

\subsection{Design of Language Instruction in \dataset{}}
For object manipulation tasks, it is essential to establish a one-to-one correspondence between actions and language instructions. Specifically, each language instruction should uniquely identify both the target object to be manipulated and the specific action to be performed. However, Ego4D's~\cite{grauman2022ego4dworld3000hours} language instructions are insufficiently detailed to satisfy this criterion.

\begin{figure*}
  \centering
  \includegraphics[width=1.0\linewidth]{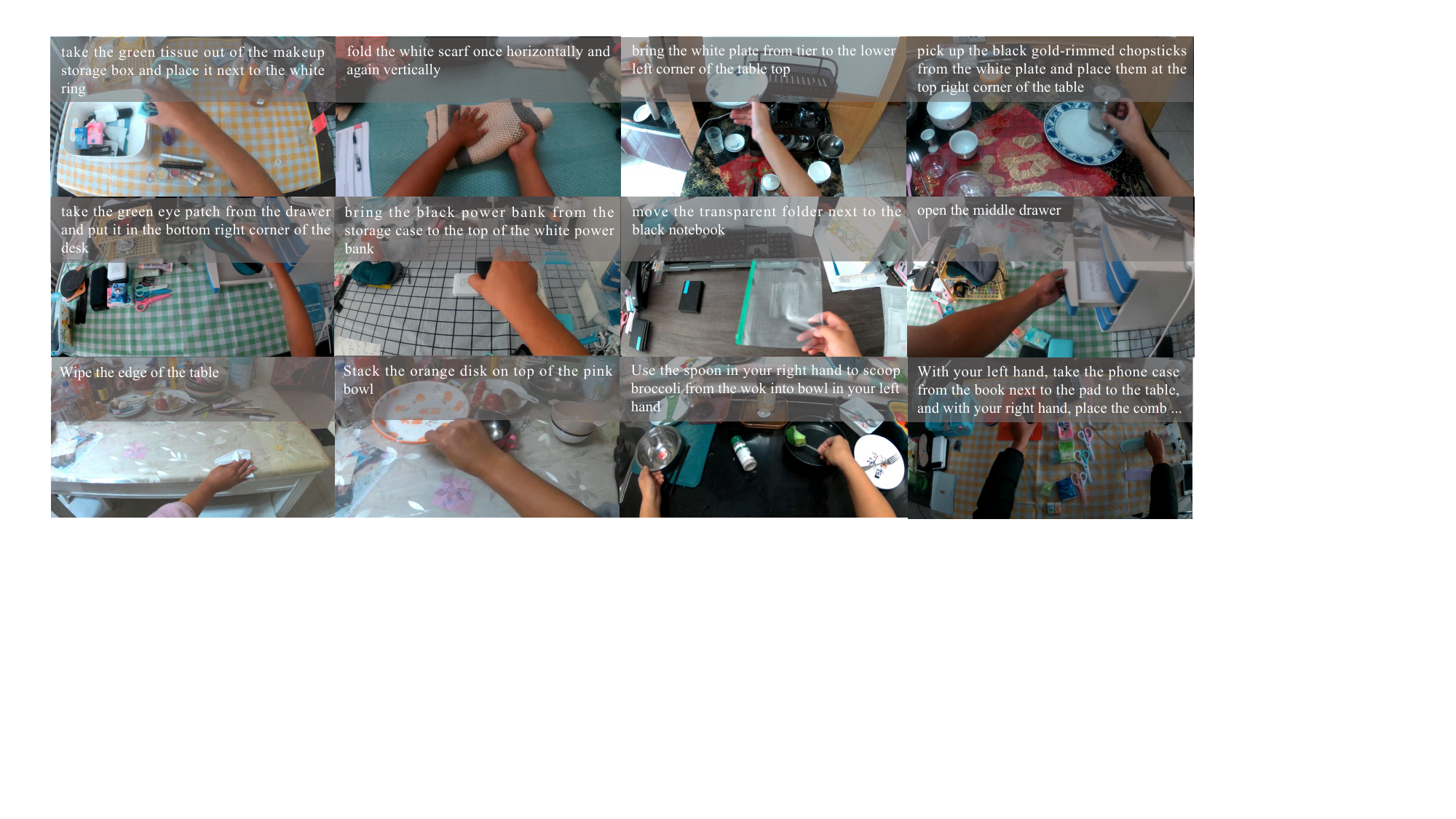}
  \caption{\textbf{Examples of video frame and corresponding language instruction.} In \dataset{}, we enhance the language instructions with more specific details to ensure accurate identification of unique target objects and their corresponding actions.}
  \label{fig:determiner}
\end{figure*}

\paragraph{Identifying Unique Target Objects.} The language instructions in Ego4D~\cite{grauman2022ego4dworld3000hours} only specify the object types without providing distinctive determiners. For instance, given the instruction ``C picks up a bowl and puts down it'', when multiple bowls are present on the table, any of them could potentially be considered as the target object. In \dataset{}, we incorporate distinctive determiners to uniquely identify objects, such as ``red bowl'', ``bowl with a tomato'', and ``bowl next to the kettle''. Figure \ref{fig:determiner} illustrates examples of video frames and their corresponding instructions from \dataset{}. 

\begin{figure}
  \centering
  \includegraphics[width=1.0\linewidth]{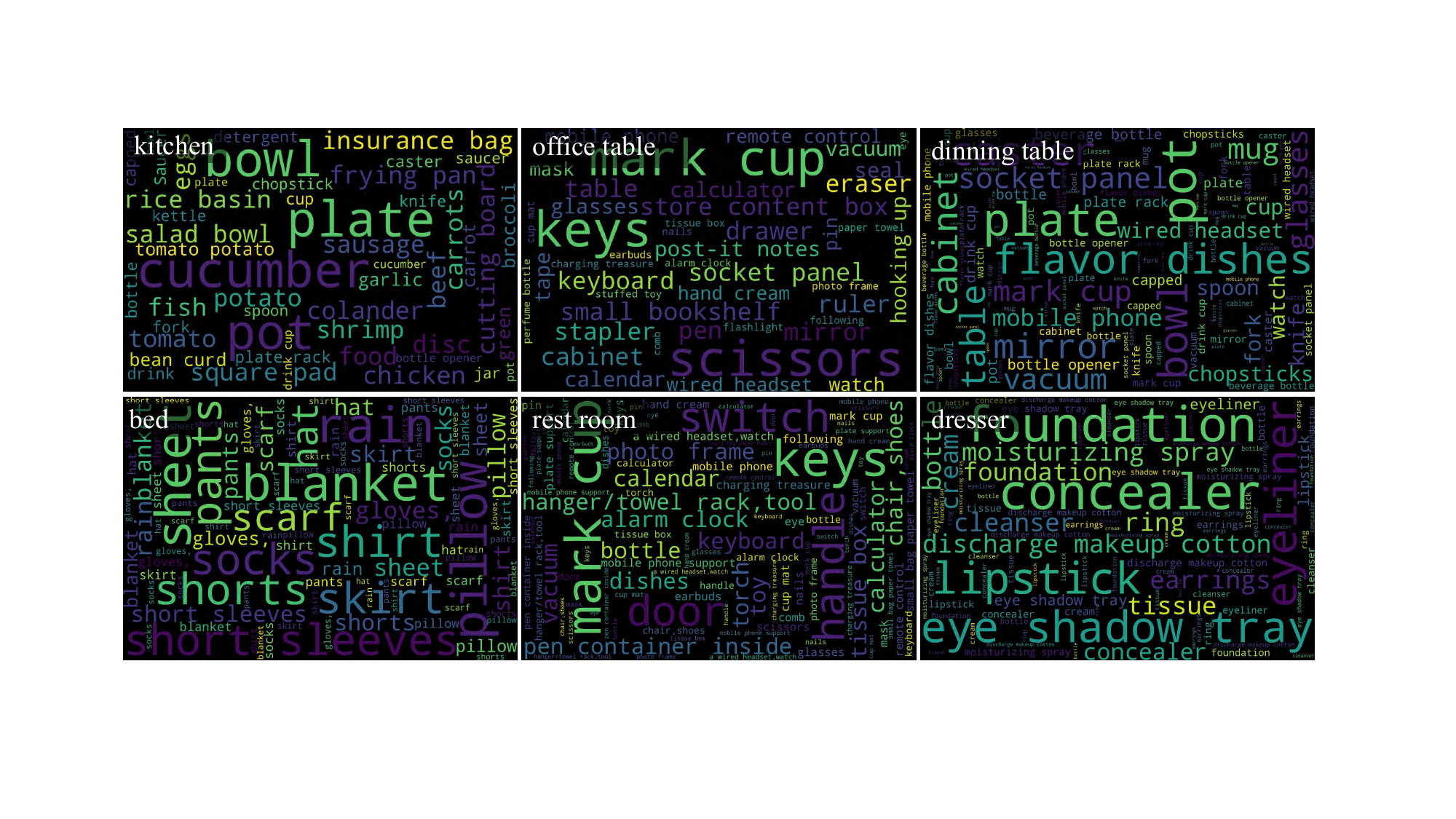}
  \vspace{-15pt}
  \caption{\textbf{Distribution of objects in videos from six scenes.} The diverse range of objects across different scenarios demonstrate the richness of activities captured in \dataset{}}
  \label{fig:wordcloud}
  \vspace{-20pt}
\end{figure}

\paragraph{Identifying Unique Actions.} Regarding object manipulation tasks, Ego4D's~\cite{grauman2022ego4dworld3000hours} language instructions merely describe the action without specifying the target location. For instance, the instruction ``C puts down the bowl'' indicates the action ``puts down'' but fails to specify the intended placement location. In \dataset{}, we provide comprehensive instructions that include detailed spatial information, such as ``take the solid black pen and put it in the pen holder'' and ``take the orange eraser and put it on the pencil case''. 

\subsection{Diversity of Manipulated Objects in \dataset{}} 
The distribution of manipulated objects in \dataset{} across different scenarios is visualized through word clouds in Figure \ref{fig:wordcloud}. These visualizations emphasize context-dependent objects, such as kitchen utensils (bowls, spoons, plates) in cooking activities and Clothing (skirts, shorts, hats) in bed environment. Additionally, some objects appear universally across scenarios, including cups, tissues, and mobile devices. The broad spectrum of objects encountered in \dataset{} reflects its comprehensive coverage of diverse activities.

\begin{figure*}
  \centering
  \includegraphics[width=1.0\linewidth]{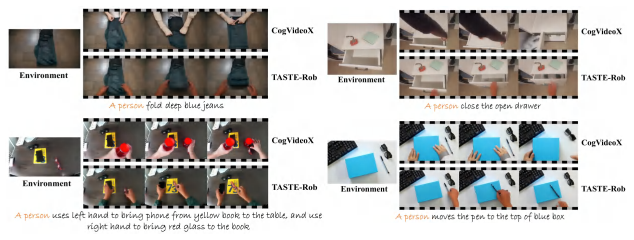}
  \caption{\textbf{More results of video generation comparison between TASTE-Rob and CogVideoX~\cite{yang2024cogvideoxtexttovideodiffusionmodels}.}}
  \label{fig:comp_appendix}
  \vspace{-15pt}
\end{figure*}

\section{Analyzing Details of hand poses}
\label{appendix:hamer}

\subsection{MANO parametric hand model}
Given a RGB image of hands, HaMeR~\cite{pavlakos2024reconstructing} is able to reconstruct 3D hand meshes by predicting MANO~\cite{Romero_2017} parameters (pose parameters and shape parameters). Specifically, the pose parameters are decomposed into two components: global orientation and hand pose. Global orientation parameters are represented by $\theta^g \in \mathbb{R}^{1 \times 3 \times 3}$, defining the overall 3D orientation of the hand in space. Hand pose parameters control the articulation of $15$ hand joints, which are represented by $\theta^p \in \mathbb{R}^{15 \times 3 \times 3}$, defining the full hand articulation. We perform a comprehensive analysis of hand pose distribution by examining both global orientation and hand pose parameters from 16 sampled frames across \dataset{} videos.

\subsection{Calculating Details of analyzing Distribution}
\paragraph{Orientation Distribution.} To analyze palm orientation, we first extract the normal vector of the palm plane from $\theta^g$. This normal vector is then projected onto the frame plane (X-Y plane), and we calculate the projected angle, which tells us how much the palm's direction deviates from the positive X-axis in the frame plane. Specifically, 0 degree points along the positive positive X-axis. We also normalize this projected angle to the 0-360 degree range. Finally, we divide the range into four 90-degree intervals to analyze the distribution of palm orientations across different directions.

\paragraph{Inter-finger Angles Distribution.} To analyze the inter-finger angles, we calculate the angles between thumb-index and index-middle finger pairs. For each frame, we first extract the normalized direction vectors of fingers ($\bm d_t$ for thumb, $\bm d_i$ for index, and $\bm d_m$ for middle finger) using their joint positions. The inter-finger angles are computed as the inverse cosine of the dot product between corresponding vectors: $\arccos(\bm d_t \cdot \bm d_i)$ and $\arccos(\bm d_i \cdot \bm d_m)$. After obtaining these angles, we plot their distributions as probability density curves, and these curves illustrate how finger spreads are distributed in \dataset{}.

\paragraph{Finger curvature Distribution.} To analyze finger curvature patterns, we compute the bending angle for each finger. Each finger has three joints: metacarpophalangeal (MCP) near the palm, proximal interphalangeal (PIP) in the middle, and distal interphalangeal (DIP) near the fingertip. We calculate the bending angle by obtaining two normalized direction vectors: one from MCP to PIP joint $\bm j_{M2P}$, and another from PIP to DIP joint $\bm j_{P2D}$. The bending angle is then computed as the inverse cosine of the dot product between these vectors: $\arccos(\bm j_{M2P} \cdot \bm j_{P2D})$. We plot these angles' distributions as probability density curves, which reveal the common curvature patterns and articulation preferences of different finger joints in \dataset{}.

\section{Our Introduced Metrics}
\label{appendix:GPV}

\begin{figure}[h]
  \centering
  \includegraphics[width=1.0\linewidth]{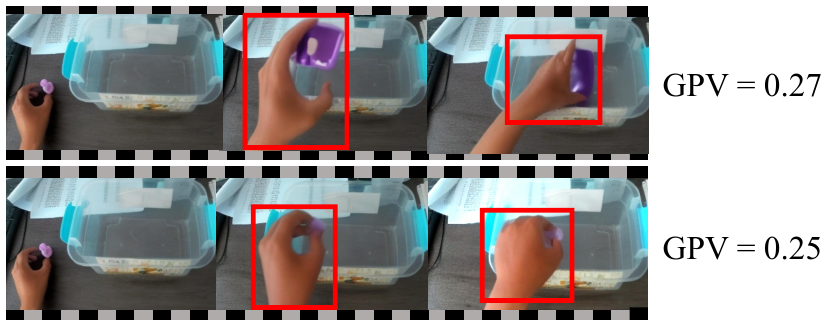}
  \caption{\textbf{Examples of generated videos and corresponding GPV value.} A higher GPV value indicates less consistency in grasping poses. 
  .}
  \vspace{-10pt}
  \label{fig:gpv}
\end{figure}

\begin{figure}
  \centering
  \includegraphics[width=1.0\linewidth]{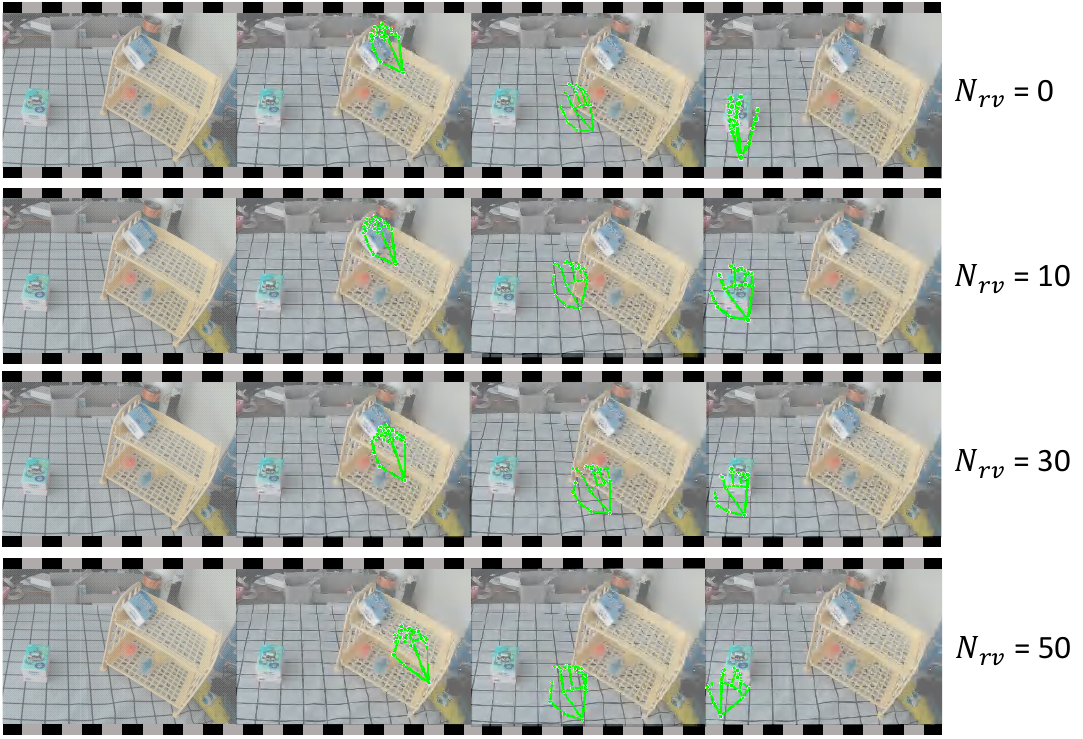}
  \caption{\textbf{Visualization of different denoising steps of pose refinement.} When $N_{rv} = 0$, no pose refinement is applied, whereas $N_{rv} = 50$ indicates that our learnable MDM generates pose sequences directly from Gaussian noise.}
  \label{fig:mdm-denoise}
  \vspace{-10pt}
\end{figure}

\paragraph{Grasping Pose Variance.} As illustrated in Section \ref{appendix:hamer}, we employ HaMeR~\cite{pavlakos2024reconstructing} to predict accurate global orientation parameters $\theta^g$ and hand pose parameters $\theta^p$. During grasping, variations in global orientation parameters are natural as hands need to move objects. However, hand pose parameters should remain consistent during grasping, as variations in grasping pose may lead to unstable grasps. To evaluate this consistency, we introduce Grasping Pose Variance (GPV), which computes the variance of hand pose parameters $\theta^p$ across frames containing grasping poses within a single video. Firstly, we utilize an off-the-shelf hand-object detector~\cite{Shan20} to extract $L^{'}$ frames containing grasping poses from a video, where the hand-object detector can detect whether hands are interacting with objects. Then, we utilize HaMeR~\cite{pavlakos2024reconstructing} to predict hand pose parameters $\{\theta^p_i\}_{i=1}^{L^{'}}$ of these frames, and calculate GPV as $\frac{1}{L^{'}}\sum_{i=1}^N(\theta^p_i-\bar{\theta^p})^2$, where $\bar{\theta^p} = \frac{1}{L^{'}}\sum_{i=1}^N \theta^p_i$. Besides, for videos containing both hands, we compute GPV values for each hand independently and take their mean value.

\begin{figure*}[h]
  \centering
  \includegraphics[width=1.0\linewidth]{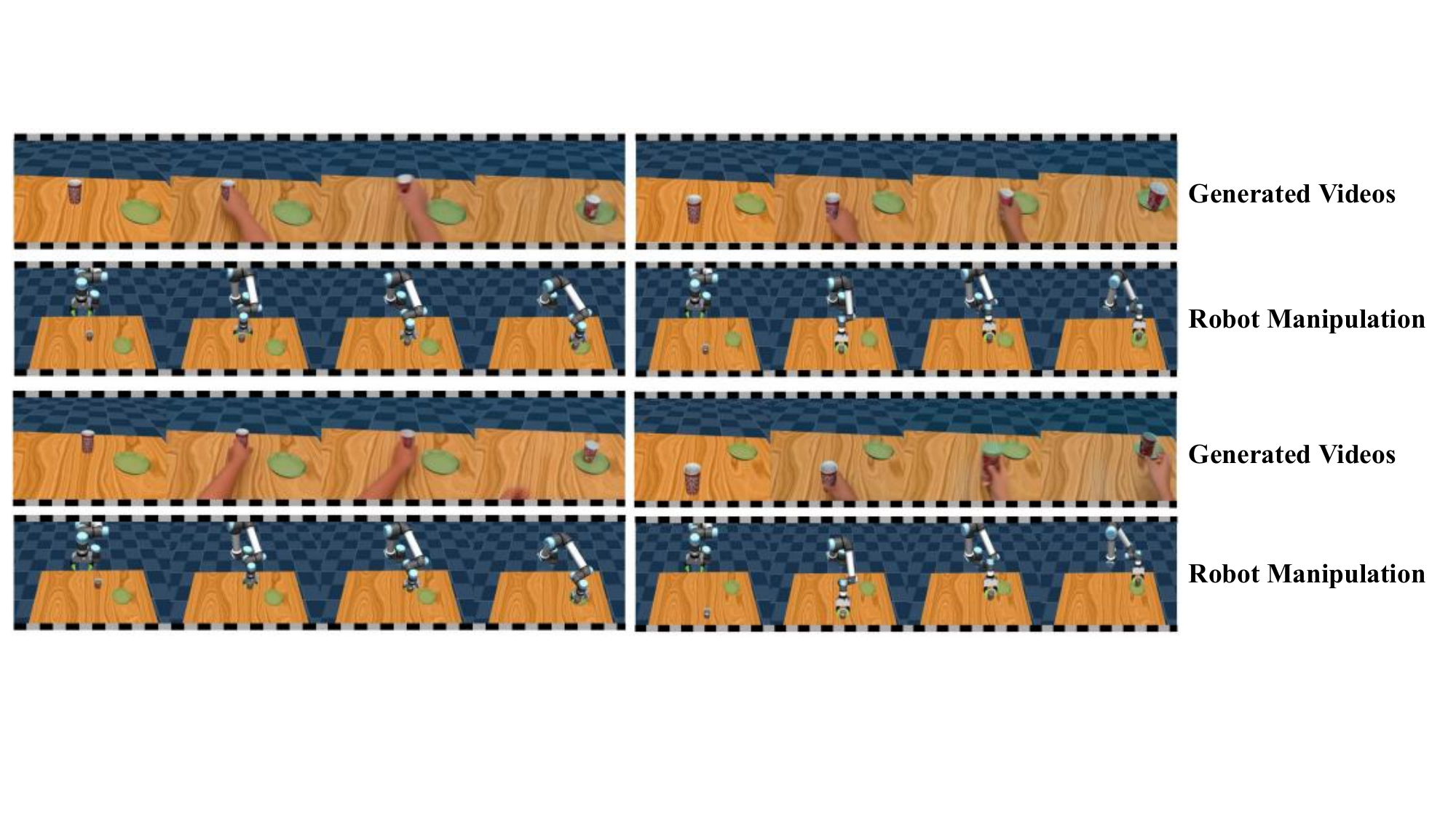}
  \caption{\textbf{Examples of video generation and robot manipulation in simulation platform.}}
  \label{fig:RM}
  \vspace{-10pt}
\end{figure*}

\paragraph{Visualization of Grasping Pose Variance.} As illustrated in Figure \ref{fig:gpv}, we compare the GPV values of two generated videos. The first case displays a video generated by our coarse action planner, while the second case presents a video generated by our complete pose-refinement pipeline. The results show that the second case maintains consistent grasping poses, whereas in the first case, the grasping poses vary significantly from pinching to holding, leading to a higher GPV value. In detail, each generated video consists of $L=16$ frames, with the corresponding grasping sequence length $L^{'}$ ranging from 6 to 8 frames. For visualization clarity, we only display a few frames in Figure \ref{fig:gpv}.

\paragraph{Grasping Type Classification Error (GTCE).} First, we extract 8 frames from both the ground truth and generated videos during the manipulation process. Then, in order to detect and classify grasping types, we use HaMeR~\cite{pavlakos2024reconstructing} to predict the corresponding pose parameters for these frames. Finally, we classify hand poses into 17 grasping types as in~\cite{7243327} and evaluate the generated poses by calculating the classification error against the ground truth.

\paragraph{Hand Movement Direction Accuracy (HMDA).} HMDA assesses the quality of generated poses by quantifying the disparity in hand movement directions between generated and ground truth videos. Initially, hand keypoints are extracted from each frame of both videos via MediaPipe~\cite{lugaresi2019mediapipeframeworkbuildingperception}. Subsequently, movement vectors are derived by computing the differences between consecutive keypoints. The angle differences between these vectors are then calculated, which effectively measures the divergence in hand movement directions. Ultimately, the metric value is obtained by computing the average of these angle differences.

\section{More Qualitative Results}
\label{appendix:result}

\begin{table}[h]
 \centering
 \begin{tabular}{@{}ccccccc@{}}
   \toprule
   \multirow{2}{*}[-0.5ex]{\centering $N_{rv}$} & \multicolumn{3}{c}{Generation Quality} \\
   \cmidrule(l){2-4}
   & KVD$\downarrow$ & FVD$\downarrow$ & PIC$\uparrow$ \\
   \midrule
   0 & 0.04 & 10.85 & 0.88 \\
   10 & \textbf{0.03} & \textbf{9.43} & \textbf{0.90} \\ 
   20 & 0.03 & 9.97 & 0.89 \\
   30 & 0.04 & 10.04 & 0.90 \\
   40 & 0.04 & 10.70 & 0.87 \\
   50 & 0.06 & 11.27 & 0.88 \\
   \bottomrule
 \end{tabular}
 \caption{\textbf{Quantitative comparison among different $N_{rv}$.} }
 \label{tab:comp_dataset}
  \vspace{-15pt}
\end{table}

\paragraph{Different Denoising Steps of Pose Refinement.} As illustrated in Figure \ref{fig:mdm-denoise}, we refine the inconsistent grasping poses while maintaining spatial awareness through a $N_{rv}$-step denoising process, starting from pose sequences extracted from our coarse generated videos ($N_{rv} = 0$). During inference, we set $N_{rv}$ as 10.

\paragraph{Comparisons on Video Generation.} As shown in Figure~\ref{fig:comp_appendix}, TASTE-Rob is able to generate high-quality HOI videos on various environment and tasks. We also provide corresponding videos in the supplementary materials.

\paragraph{Results on Robot Manipulation.} As shown in Figure \ref{fig:RM}, TASTE-Rob is able to generate high-quality HOI videos on robot simulation platform and achieve accurate robot manipulation.

\section{More Implementation Details}
\label{appendix:implementation}

\subsection{Stage I: A Coarse Action Planner}
In Stage I, we fine-tune DynamiCrafter~\cite{xing2023dynamicrafteranimatingopendomainimages} on \dataset{} from its released checkpoint at 512 resolution. The fine-tuning process is conducted with a total batch size of 16 and a learning rate of $5 \times 10^{-5}$ for 30K steps. For inference, we use a 50-step DDIM sampler and adopt the same multi-condition classifier-free guidance as in DynamiCrafter~\cite{xing2023dynamicrafteranimatingopendomainimages}.

\paragraph{Video Sampling Strategy.} Before fine-tuning, existing general I2V latent diffusion models~\cite{xing2023dynamicrafteranimatingopendomainimages,yang2024cogvideoxtexttovideodiffusionmodels,zhang2023i2vgenxlhighqualityimagetovideosynthesis,wang2023videocomposercompositionalvideosynthesis} obtain video frames $\bm v$ by sampling from a short clip within a longer video, with limited focus on maintaining action integrity in the generated videos.
However, our HOI video generation requires strict action integrity, as it directly impacts the effectiveness of robot manipulation learning. Therefore, we sample frames from the entire video to ensure complete action preservation. For clarity, we denote $\bm v$ as $\bm v^{0:L-1}$, and $\bm v^i \in \mathbb{R}^{3 \times H \times W}$ as the $i$-th frame of $\bm v$. In our work, we set $\bm{v}^{0}$ as the start frame of the entire video, $\bm{v}^{L-1}$ as the end frame, and obtain the intermediate frames $\{\bm{v}^{1}, \bm{v}^{2}, ..., \bm{v}^{L-2}\}$ through uniform sampling across the entire video, ensuring continuity and completeness in action representation. Specifically, we set $L$ as 16, and generate videos consisting of 16 frames.

\subsection{Stage II: Revising Hand Pose Sequences}
In Stage II, we train our learnable MDM~\cite{tevet2023human} on TASTE-Rob. We set the 60-frame hand pose key-points normalized coordinates (extracted from videos by MediaPipe~\cite{lugaresi2019mediapipeframeworkbuildingperception}) as motion parameters and train MDM to generate these coordinates conditioned on environment images and language instructions. The training is conducted with a batch size of 64 and a learning rate of $1 \times 10^{-4}$ for 500K steps. During inference, MDM employs a 50-step DDIM sampler when generating from Gaussian noise. For pose refinement, we apply a 10-step denoising process from the extracted coarse hand pose sequences and adopt the same classifier-free guidance as in~\cite{tevet2023human}.

\subsection{Stage III: Regenerating with Refined Pose}
In Stage III, we fine-tune our learnable frame-wise adapter from Stable Diffusion 2. The training is conducted with a batch size of 32 and a learning rate of $5 \times 10^{-5}$ for 30K steps.
For training, the frame-wise adapter take the 16-frame visualized hand pose images as input. Specifically, we extract the 16-frame hand pose key-point coordinates and then draw their single-channel visualized images. For inference, we first draw visualized images from the refined hand pose sequences. Then, we generate videos conditioned on these visualized images, environment images and language instructions, applying a 50-step denoising process and the same multi-condition classifier-free guidance as ToonCrafter~\cite{xing2024tooncrafter}.


\end{document}